\documentclass[3p,times]{elsarticle}

\usepackage{amssymb}
\usepackage{amsmath, graphicx, xfrac, amsfonts, amsbsy, algorithm, color, subfig, url, multirow, booktabs, caption}
\usepackage[noend]{algpseudocode}
\usepackage[figuresright]{rotating}
\def\R {\mathbf{R}}

\def\t {\mathbf{t}}

\def\A {\mathbf{A}}
\def\x {\mathbf{x}}

\def\X {\mathbf{X}}
\def\tX {\tilde{\X}}
\def\Y {\mathbf{Y}}

\def \SO3{ \mathbf{SO}(3) } 
\def \SE3{ \mathbf{SE}(3) } 
\def \Rspace { \mathbb{R}}

\def\argmin{\mathop{\mathrm{arg\,min}}}

\makeatletter
\def\BState{\State\hskip-\ALG@thistlm}
\makeatother

\begin{document}

\begin{frontmatter}



%% e.g. \dochead{17th International Conference on Dynamical Processes in Excited States of Solids}

\title{ForestAlign: Automatic Forest Structure-based Alignment for Multi-view TLS and ALS Point Clouds.}

%ForestAlign: Automatic Forest Structure-based Alignment for Multi-view TLS and ALS Point Clouds.

\author[lanl]{ Juan Castorena }
\author[lanl]{ L. Turin Dickman}
\author[lanl]{ Adam J. Killebrew}
\author[lanl]{James R Gattiker}
\author[lanl]{Rod Linn}
%\author[srs]{Scott Pokswinski}
\author[srs]{E. Louise Loudermilk}

\address[lanl]{Los Alamos National Laboratory, Los Alamos, NM, 48124 USA}
\address[srs]{Southern Research Station, 200 W.T. Weaver Blvd. Asheville, NC 28804-3454, USA}

\begin{abstract}
Access to highly detailed models of heterogeneous forests, spanning from the near surface to above the tree canopy at varying scales, is increasingly in demand. This enables advanced computational tools for analysis, planning, and ecosystem management. LiDAR sensors, available through terrestrial (TLS) and aerial (ALS) scanning platforms, have become established as primary technologies for forest monitoring due to their capability to rapidly collect precise 3D structural information directly. Selection of these platforms typically depends on the scales (tree-level, plot, regional) required for observational or intervention studies. Forestry now recognizes the benefits of a multi-scale approach, leveraging the strengths of each platform while minimizing individual source uncertainties. However, effective integration of these LiDAR sources relies heavily on efficient multi-scale, multi-view co-registration or point-cloud alignment methods. In GPS-denied areas, forestry has traditionally relied on target-based co-registration methods (e.g., reflective or marked trees), which are impractical at scale. Here, we propose ForestAlign: an effective, target-less, and fully automatic co-registration method for aligning forest point clouds collected from multi-view, multi-scale LiDAR sources. Our co-registration approach employs an incremental alignment strategy, grouping and aggregating 3D points based on increasing levels of structural complexity. This strategy aligns 3D points from less complex (e.g., ground surface) to more complex structures (e.g., tree trunks/branches, foliage) sequentially, refining alignment iteratively. Empirical evidence demonstrates the method's effectiveness in aligning TLS-to-TLS and TLS-to-ALS scans locally, across various ecosystem conditions, including pre/post fire treatment effects. In TLS-to-TLS scenarios, parameter RMSE errors were less than 0.75 degrees in rotation and 5.5 cm in translation. For TLS-to-ALS, corresponding errors were less than 0.8 degrees and 8 cm, respectively.
These results, show that our ForestAlign method is effective for co-registering both TLS-to-TLS and TLS-to-ALS in such forest environments, without relying on targets, while achieving high performance.
\end{abstract}


\end{frontmatter}

%%%%%%%%%%%%%%%%%%%%%%%%%%%%%%%%%%%%%%%%%%%%%
%% Sections
%%%%%%%%%%%%%%%%%%%%%%%%%%%%%%%%%%%%%%%%%%%%%
%\linenumbers

\section{Introduction}
\label{Sec:intro} 

% Motivation
%Forests cover approximately four billion hectares amounting to roughly $~31 \%$ of the earth's land area \cite{FaoUnep2020}. %In 2020, the rate of tree cover loss due to fire only is increasing to $~4 \%$ every year \cite{FaoUnep2020}. 
%Demand for tools that help maintain the balance of a healthy forest ecosystem are increasing but their complexity is challenging. %as it depends on a wide range of factors including resilience against disease and fire, health and biodiversity \cite{white2016remote}. 
Forest monitoring methods that consider the spatio-temporal heterogeneity of the forest are increasing, driven by demands for improved planning, management, analysis, and more effective and efficient decision-making \cite{Atchley2021}.
%Methods that help collect information about these factors and that consider the heterogeneity of the forest and its changes over time is an active area of research seeking to provide the means for improved planning, management, analysis and more effective and efficient decision-making \cite{Atchley2021}. 
Traditionally, one of the most widely used methods of forest characterization is standard field plot surveys employing spatial sampling and estimation techniques. These are used to quantify forest factors including growing stock volume, biomass, carbon balance, and tree measurements (e.g., canopy cover, diameter at breast height, crown width, height) \cite{tomppo2010national}. However, these plots are typically sampled manually in the field, which can be economically costly and time-consuming, especially when considering large spatial scales (e.g., 1 million ha).
\begin{figure} [ht]
	\centering 
	\includegraphics[width=0.6\linewidth]{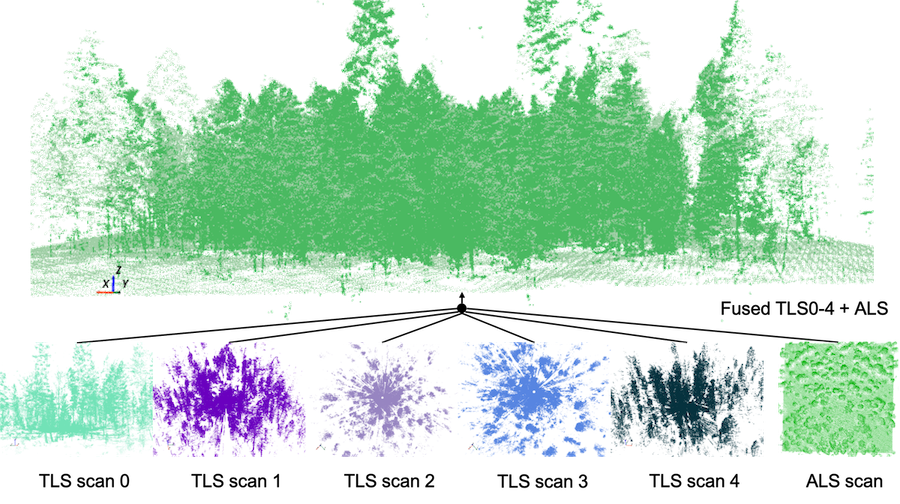}%multiview_side_2}}
	\caption{Alignment and accumulation of 5 multi-view TLS and ALS in New Mexico forest.}
	\label{fig:align_result} 
\end{figure}
%
%
% Remote sensing 
Remote sensing techniques, including terrestrial laser scanning (TLS), aerial laser scanning (ALS), and optical photogrammetry, have been established as technologies offering time and cost economies at scale in forest surveys \cite{white2016remote}. Among these, LiDAR-based scanning has gained widespread interest due to its inherent capability to rapidly collect precise 3D structural information over large regional extents \cite{dubayah2000lidar}.
% A few sentences describing advantages and limitations of ALS and TLS.
Airborne based LiDAR for example, utilizing accurate position sensors including RTK, enables data collection at the regional/landscape scale. %at spatial resolutions in the order of $\sim$10 pts/m$^2$. 
It has been shown to be effective for forest characterization and prediction of forest factor metrics \cite{hyyppa2012advances}. However, some limitations exist as it has difficulty penetrating the tree canopy cover \cite{nelson1984determining}, resulting in low-resolution spatial variability along the tree stand direction, even with its powerful but computationally demanding full-waveform (FW) based technology.
%It has shown to be effective for prediction of forest factors \cite{hyyppa2012advances}, however, some limitations exist as it yields low-resolution spatial variability and has problems penetrating the tree canopy cover in dense areas even with its powerful but computationally demanding full-waveform (FW) based technology.
\textit{In-situ} TLS, on the other hand, is complementary to ALS as it is capable of collecting information at a much higher 3D spatial resolution along the full vertical tree stand direction (i.e., ground surface, mid-tree story, and canopy vegetation) \cite{hilker2010comparing}. However, TLS is limited to a few tens of meters in range, and scaling its advantages over larger area extents requires more sophisticated methods of sensor placement stratification and data source fusion. %strategies and methods for moving the scanning system and fusing their data from multiple scans to increase the field of view can be challenging. The simplest of strategies consists of placing the scanner at different locations within a plot (i.e., multi-view TLS), but other more sophisticated options exist including hand-held scanners \cite{holopainen2013tree}. 
Having access to a diversity of sensing sources and a proper way to combine them can enhance capabilities, enabling precise 3D information from near-surface vegetation to the top of the canopy at a characteristic scale rarely available when a single scanning source is used \cite{hilker2010comparing}. Here, we focus on this problem and propose a new 3D point cloud co-registration method suitable for forest environments, with the purpose of combining multi-view, multi-source LiDAR data. %to enable the enhanced capabilities of an integrated sensing source framework for advanced forest monitoring. 
The proposed method is fully automatic, does not require any targets, and does not rely on the presence of man-made objects in the scene, as is typically required by conventional point cloud co-registration approaches in forestry.
%Here, we focus our scope on one of the fundamental enablers for LiDAR data integration combining multi-view multi-source LiDAR data in forest ecosystems: co-registration or alignment of LiDAR source data.

%Relevant work dealing with the alignment of multi-view LiDAR sources, is a well known problem in urban settings involving man-made objects. 
% TLS-to-TLS in urban scenes 
The problem of multi-view LiDAR source alignment is well known, with highly efficient and effective methods available for point clouds collected from urban scenes and/or highly structured man-made objects. In the case of TLS-to-TLS co-registration, for example, the work of \cite{brenner2008coarse} approximated scenes by planar surfaces and used them as features to globally optimize the alignment parameters. In \cite{sumi2018multiple}, the method projects 3D points into a ground occupancy 2D plane image and co-registers based on these occupancy plane representations. Along a similar principle, \cite{xiong2022fast} proposed a bird's eye view 3D-to-2D point cloud projection method and used SIFT features to register scan pairs.
% TLS-to-ALS in urban
In the multi-scale case of TLS-to-ALS co-registration, the work of \cite{kedzierski2014terrestrial} used wavelets to detect building edges and subsequently matched these between TLS and ALS scans for alignment. In \cite{teo2014surface}, they matched planar structures instead, while \cite{wu2014feature} used both edges and planes to register scans containing buildings. The works of \cite{yang2015automated, cheng2018automatic, ling2022graph} extracted urban building outlines, contours, and boundaries and their correspondences between scans, thus depending on the regularity of corners, edges, and planes.
An important aspect to highlight in the aforementioned works is that, although they provide scalable approaches by being both fully automatic and without the need for manual target placements, these \textit{do not work} in the more complex case of point clouds collected from forest ecosystems, where the presence of highly structured features (e.g., edges, lines, planes) is limited or even absent.

% Problems with existing LiDAR scan registration methods in forestry
For these reasons, most studies exploiting combined multi-view TLS and ALS in forestry \cite{hilker2012simple, holopainen2013tree, kankare2014accuracy, bauwens2016forest, ge2021target} have utilized either known physically placed reference targets (e.g., multiple uniform-size spheres, retro-reflective targets), which are detected and matched between scans, or have manually extracted 3D point correspondences for alignment. Unfortunately, both of these methods tend to render multi-source LiDAR integration impractical for large-scale applications. In addition, target-based methods are not suitable or are practically challenging to deploy in cases involving TLS-to-ALS co-registration, as time disparities between TLS and ALS data typically exist, and occlusion from trees can become a challenge for target visibility from both terrestrial and aerial perspectives.

Target-less methods in forestry, on the other hand, are known to be less time-consuming and offer the possibility to scale to large area extents and to many scans. For example, the works of \cite{bienert2009methods, kelbe2016marker, liu2017automated, tremblay2018towards, polewski2019marker, guan2019novel, dai2020fast, wang2021efficient, ge2021global} use semantic segmentation to extract tree features (e.g., stem size, stem axis, breast height, distances between trees) and find tree correspondences between the LiDAR scan sources.
% TLS2TLS in forestry
Notable examples in the TLS-to-TLS case includes the recent work by \cite{xu2022automatic}. Their method relies on semantic segmentation of tree stem and branch centers and finding correspondences between scans, but it is restricted to single tree scans. 
%The work of \cite{guan2020marker}, on the other hand, used the starting points of tree LiDAR shadows (i.e., vertex of cone shaped projection areas without LiDAR returns caused by tree occlusion) as features to co-register adjacent TLS scans. Accuracies at the level of target based methods were achieved, but the evaluation performed in the study was limited to only 17 scan pairs. 
Other recent TLS-to-TLS methods rely on finding correspondences between extracted tree attributes, such as stems \cite{kelbe2016marker, liu2017automated, tremblay2018towards, dai2020fast} and tree locations \cite{guan2019novel}. %Their accuracies are highly dependent on the performance of semantic segmentation of tree attributes and is thus prone to co-registration errors in cases of low semantic segmentation performance which commonly occurs in high tree density forests.
% TLS2ALS in forestry
%But as in all the aforementioned methods, these works focus on registering urban scenes and their fundamental step of feature extraction/correspondence is not suitable for forest ecosystems. Recent efforts in \cite{tremblay2018towards, polewski2019marker, guan2019novel, dai2020fast, wang2021efficient} proposed methods to register LiDAR scans coming from forest environments by finding relative tree attributes. 
In the TLS-to-ALS case, works such as \cite{hauglin2014geo, polewski2019marker} find correspondences between scans through tree position extractions. The works of \cite{paris2017novel, dai2019automated} rely instead on the correspondence and alignment of tree crown features. Alignments based on relative tree heights between scan sources have also been exploited in \cite{liu2021target}. In general, the efficacy of these TLS-to-TLS and TLS-to-ALS manually selected tree feature methods depends on the accuracy of the tree segmentation step. This can become problematic with increasing tree density while also becoming increasingly time-consuming and computationally demanding. Additionally, these tree segmentation-based approaches are \textit{not applicable} in areas where there are no trees (e.g., the Savannah or thinned areas).
%Methods using manually placed reference targets do not apply in the TLS-to-ALS setting due to practical challenges including finding targets visible to both sources, and time disparities between TLS and ALS data collection surveys.
% Using position-based methods
%These methods tend, however, to be sensitive to the accuracy of the tree extraction method; a process that can become computationally demanding in dense forest areas and prone to significant errors. 
Other less computationally demanding method alternatives have resorted to position sensors, including GNSS/IMU systems \cite{holopainen2013tree}, but have found inferior performance due to higher uncertainties or missing measurements in areas of poor or nonexistent satellite-based geo-referencing \cite{liang2014possibilities}. This is typical, especially in TLS collections of densely vegetated regions.
In light of this discussion, we propose a new method termed automatic forest structure-based alignment (ForestAlign) to co-register point clouds from forest ecosystems that is fully automatic and does not require any manual target placements. The proposed method finds the co-registration parameters by using a notion of 3D structural complexity that encodes information compactness or compression about the point cloud composition. Here, we use a dense 3D plane approximation to the point cloud and a mixture of von-Mises Fisher (vMF) distributions to represent and extract the point cloud composition and the complexity or informativeness of its components.
We show that grouping and matching LiDAR scan point subsets based on this notion of 3D structural complexity is an effective strategy for co-registering pairs of scans from highly complex and heterogeneous forest environments and under a variety of conditions. The intuition behind our incremental co-registration strategy relies on the assumption that forests are composed of underlying 3D structures of varying complexity (e.g., ground, tree trunks/branches, foliage), and designing alignment approaches should exploit those that are less complex (e.g., ground, tree trunks) to find a rough alignment first, and then use those of increasing complexity (e.g., foliage) as a refinement.
Figure \ref{fig:align_result} illustrates a preliminary example aligning five plot-scale, multi-view TLS scans and a regional ALS scan collected from a New Mexico, USA forest. Advantages of the proposed method include its effectiveness in registering LiDAR scans in complex forest environments without requiring any position sensors or reference targets, explicit extraction of trees or other semantic features, while also being fully automatic and scalable to large area extents with the ability to co-register many TLS-to-TLS and TLS-to-ALS scans. In the following, Section \ref{Sec:method} describes our approach, while Section \ref{Sec:experimentation} empirically demonstrates the overall performance of our method. Finally, Section \ref{Sec:discussion} includes a short discussion about the benefits of the proposed approach in the context of forestry, and Section \ref{Sec:conclusion} summarizes our contributions.

\section{Approach}
\label{Sec:method} 

%%%%%%%%%%%%%%%%%%%%%%%%%%%%%%%%%%%
% Solution
%%%%%%%%%%%%%%%%%%%%%%%%%%%%%%%%%%%

Co-registration seeks to align available point clouds into a global, consistent coordinate system. These point clouds can originate from various perspectives, sensors, and scanning platforms (e.g., \textit{in-situ} TLS, mobile TLS, UAVs, ALS). Our focus is on methods that are automatic, targetless, and do not use position and orientation sensing. Therefore, we restrict our approach to point clouds with at least partial field of view (FOV) overlap or spatial overlap. To the best of our knowledge, the proposed approach is the first fully automatic and targetless method capable of co-registering both TLS-to-TLS and TLS-to-ALS scans under diverse forest ecosystem conditions. It does not depend on the accuracy of any prior semantic segmentation approach (e.g., tree detection and segmentation).

Here, we first include some notations used throughout this paper, along with a few definitions. A LiDAR scan point cloud is denoted by $\X \in \R^{N \times 3}$ with $N$ being the number of 3D points in the scan and 3 being the point coordinates in the $x,y$ and $z$-axes. A subscript $s$ or $t$ attached to the right of a scan as $\X_s$ is used to distinguish between a source $s$ and a target $t$ point cloud scan while a superscript $i$ as in $\X^{i}$ denotes a selection of point(s) according to index or index set $i$. Labels $\Y \in \{1,2...,K\}^N$, are group labels that indicate the group to which each 3D point belongs to. For a pair of scans, co-registration is characterized  by an aligning rigid body transformation consisting of 6 orientation/position parameters: three rotation angles ($\theta_{\text{roll}}, \theta_{\text{pitch}}, \theta_{\text{yaw}}$) and three translations ($t_x,t_y,t_z$). 3D rotations are represented by the matrix $\R \in \SO3$ in the space of the special orthogonal group $\SO3$ and fully expressed by $\theta_{\text{roll}}, \theta_{\text{pitch}}, \theta_{\text{yaw}}$, while translations are equivalently denoted as the vector $\t \in \Rspace^3$ of $t_x,t_y,t_z$ coordinate elements. %The subscript and superscripts attached to the right of $\R, \t$ denotes that this transformation takes from the coordinate system of the subscript, to that of the superscript while 
The superscript $k$ attached to the right of  $\R, \t$ denotes the $k$-th estimate.
\begin{figure} [ht]
	\centering 
	\includegraphics[width=1.0\linewidth]{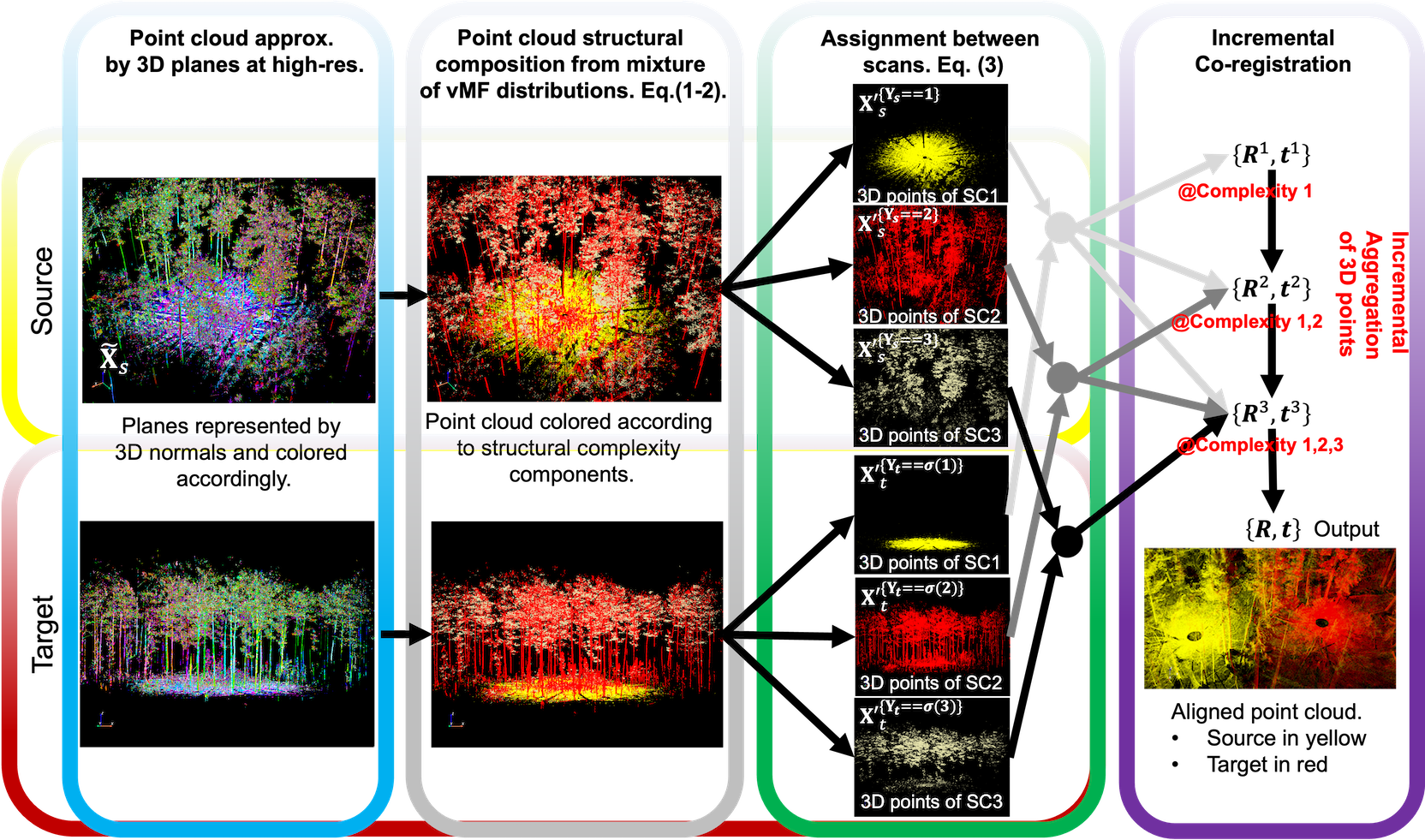}
	\caption{Co-registration schematic. Two point-clouds, labeled source and target, are co-registered using our 3D structural complexity-based approach. On the left, point clouds are color-coded according to the parameters of a 3D plane approximation. Note that colors in the ground and tree trunks/branches appear more uniform compared to foliage, where colors are more randomly distributed or noisy. In the second column, we use a mixture of von Mises-Fisher (vMF) distributions (with the number of groups $K=3$) to group points according to their structural complexity, compactness, or informativeness. In the third column, we split the point-clouds according to the structural groups and assign a correspondence between scan groups. Finally, the right-most block co-registers in an incremental fashion by aggregating 3D points of increasing structural complexity, refining the estimated co-registration parameters. %The overall approach leverages the property that points belonging to forest components of lower structural complexity (e.g., ground, tree trunks/branches) are more uniformly distributed, whereas those in higher complexity components (e.g., foliage) are more randomly distributed. This natural grouping aids in effectively aligning the point clouds.
    }
	\label{fig:schematic} 
\end{figure}

\subsection{ Pair-wise co-registration} \label{Ssec:registration}
% Main idea of the method
%The method of registration takes overlapping pair-wise point clouds and finds the relative 6 degrees of freedom (DOF) consisting of three (roll, pitch and yaw) rotations and three translations ($x,y,z$) characterizing their alignment. The method can handle limiting overlapping cases and aligns pair-wise point clouds locally (i.e., in the vicinity of the true alignment transformation). 
The proposed co-registration approach aligns pairs of scans by incrementally aligning matching groups of 3D point subsets grouped and matched by structural complexity similarities. Our approach groups and incrementally aggregates 3D points of increasing structural complexity, thus sequentially refining its previous alignment estimates. The intuition behind such a strategy is that 3D points belonging to components with less structural complexity (e.g., ground, tree trunks) should be used first to find a rough alignment between point-clouds, and then the 3D points of higher structural complexity are used to refine the alignment in subsequent steps. The motivation for this incremental structural complexity strategy is based on our empirical findings showing that high complexity forest components (e.g., foliage) can cause co-registration approaches to become stuck at an incorrect alignment. Unlike automatic co-registration approaches in forestry that use forest semantics, our method utilizes a notion of 3D structural complexity without explicitly caring about semantics. 
%This approach is less computationally intensive and avoids the additional errors introduced by tree segmentation methods. 
Additionally, our method differs from standard co-registration methods outside of forestry, which typically rely on features characteristic of human-made objects or scenes (e.g., corners, edges, planar facets, spheres), which rarely appear in forest environments.

The main ingredients of our co-registration approach include: (1) metrics to quantify or represent the 3D structural complexity of point-cloud components, implicitly modeling the underlying forest composition; (2) the process for grouping and matching 3D points into levels of structural complexity; and (3) the optimization strategy to find the co-registration parameters based on the 3D points grouped at different levels of structural complexity. In the following sections, we present the mathematical descriptions and steps encompassing each of these components, while an overall schematic summary of the proposed approach is provided in Fig.\ref{fig:schematic}.

\subsubsection{3D structural complexity} \label{Ssec:similarity_metrics}
Given a point cloud $\X$, we propose to quantify the 3D structural complexity of the underlying forest components (e.g., ground, tree trunks/branches, foliage) by the level of information compactness or compression possible in a distribution of 3D plane approximations to the point cloud. The first stage approximates a point cloud $\X$ by the same number of 3D planes as there are 3D points in $\X$. The second stage models the distribution of 3D planes as a mixture of $K$ distributions, providing a natural grouping of 3D planes into $K$-levels, each characterized by a distinct variability or compactness.
A 3D plane is computed at each point using the 3D points in a spherical neighborhood of radius $0.25$(m) centered at the 3D point of interest and by computing the eigenvalue/vector decomposition. 
%Given a neighborhood of 3D points, we extract the 3D plane through the random sampling and consensus (RANSAC) algorithm \cite{fischler1981random}, which is known to alleviate fitting in the presence of outliers. In this case, however, we use it to alleviate the effect of grass, small shrubs, leaves, debris, and of other factors that make forests, structurally more complex. 
3D planes are represented here by their 3D normal vectors $\tilde{\X} \in \Rspace^{N \times 3}$, each of unit magnitude (i.e., $\|\tilde{\X}^{i}\|_{\ell_2} = 1$), and we summarize the statistics over 3D normals computed overall points in the point-cloud.
The distribution over normal vectors is parametrized here as a mixture of von Mises-Fisher distributions (vMF) \cite{banerjee2005clustering}. The family of vMF distributions is well suited for distributions over normal vectors, or in general, to data that represents directions with no significance on magnitudes. This distribution is fully parametrized by its directional mean $\pmb{\mu}$, describing the mean direction of normals, and a concentration parameter $\pmb{\kappa}$, describing how weak or strong the symmetric variations from the mean are. Mathematically, its probability density function is given as:
\begin{equation}
    p(\tilde{\X}^{(i)} | ~ \pmb{\mu},\pmb{\kappa}) = c_d(\pmb{\kappa}) \exp(\pmb{\kappa} \pmb{\mu}^T \tilde{\X}^{(i)} ) \quad \text{with} \quad c_d(\pmb{\kappa}) = \frac{\pmb{\kappa}^{d/2-1}}{(2 \pi)^{d/2} I_{d/2-1}(\pmb{\kappa})}
\end{equation}
where $I_{d}$ represents the modified Bessel function of the first kind and order $d$, with 3D normal dimension $d=3$. More details on the properties of the vMF distribution can be found on \cite{banerjee2005clustering}.
A small $\pmb{\kappa}$ represents small variations, while a large one represents high variations or, alternatively, a heavy-tailed distribution. In the limit, as $\pmb{\kappa} \rightarrow 0$, the vMF distribution approximates a uniform distribution, while as $\pmb{\kappa} \rightarrow \infty$, it tends to a point distribution. 
This flexibility gives the mixture of vMF distributions the ability to group 3D points into either compact or more spread clusters without assuming independent and identically distributed (iid) sample 3D point data. This property is especially helpful given the characteristic non-uniform 3D point density sampling of LiDAR, which can bias standard clustering approaches to incorrect parameter estimates as they typically enforce groups of approximately equal variance.

In the context of our application, the mixture of vMF distributions is aimed at expressing the underlying forest composition variations in 3D structural complexity. Components of low complexity are expected to have high concentration $\pmb{\kappa}$, while components of high complexity are expected to have low concentration. This is the key interpretation of structural complexity here: components of low structural complexity can be compactly represented, while components of high structural complexity have more randomness or variability, are less structured, and can be represented less compactly. Alternatively, this can also be interpreted from an information-theoretic standpoint using Shannon entropy \cite{shannon1948mathematical}. Forest components of high structural complexity are expected to be low in informativeness or entropy, and those of low complexity are expected to be high in informativeness. Under this interpretation, we can formulate the problem of computing structural complexity (SC) for a point cloud $\X$, given $K$ number of complexity groups, as:
\begin{equation} \label{SC}
    \text{SC}(\X) = \sum_{k=1}^K \text{SC}(\tilde{\X},k) = \sum_{k=1}^K H(\tilde{\X}^{\{\Y == k\}} | ~ \pmb{\mu}_k,\pmb{\kappa}_k) = \sum_{k=1}^K \sum_{i \in \{ \Y == k\}} p \left ( \tilde{\X}^{i} | ~ \pmb{\mu}_k,\pmb{\kappa}_k  \right ) \log p \left ( \tilde{\X}^{i} | ~ \pmb{\mu}_k,\pmb{\kappa}_k  \right )
\end{equation}
where the first summation is over the number of $K$ complexity groups, $H(\tilde{\X}^{\{\Y == k\}}| ~ \pmb{\mu}_k,\pmb{\kappa}_k)$ is the conditional entropy of the $k^{\text{th}}$ vMF component given parameters $\pmb{\mu}_k, \pmb{\kappa}_k$, and the second summation on the right hand side is over the 3D planes associated with each group $k$. In other words, the overall structural complexity of a point cloud $\X$ can be described by the summation over the individual component complexities $\text{SC}(\tilde{\X},k)$.
%The intuition behind Eq.\eqref{SC} is that 3D points corresponding to different components of the forest exhibit varying levels of information compactness or concentration. 
For instance, points on the forest ground surface are expected to have high concentration or information compactness in their 3D plane approximations, indicating low structural complexity. On the other hand, points representing foliage are expected to have low concentration and more variability in their 3D plane orientations, suggesting high structural complexity. It is worth noting that notions of structural scene complexity have been utilized previously (e.g., in \cite{castorena2010using, castorena2010modeling}) to sub-sample scenes at different complexity scales based on waveform envelope shapes in full-waveform LiDAR data. 

\subsubsection{Grouping and Matching} \label{Ssec:grouping}
Grouping and matching is an approach developed here to group and match point-cloud subsets using the structural complexity metrics described in Sec. \ref{Ssec:similarity_metrics}. It involves two main tasks: (1) Intra-scan groupings, conglomerating 3D point subsets of a point-cloud into levels of structural similarity and, (2) Inter-scan matching, matching structurally similar groups of 3D points between pairs of source and target point clouds. 
% Intra-scan grouping
Intra-scan 3D point group subsets are groups of 3D points in a single point cloud that belong to the same structural complexity (i.e., they have the same structural characteristics).
These groups are naturally formed using the mixture of von Mises-Fisher (vMF) distributions. To achieve this, we estimate parameters $\pmb{\kappa}_k, \pmb{\mu}_k$ for $k \in \{1,2,..,K\}$ and the labels $\Y$. The estimation process follows the approach outlined in \cite{banerjee2005clustering}, employing the expectation-maximization (EM) algorithm for parameter estimation.
The number of groups $K$ representing structural complexity can vary depending on the scene characteristics, such as the presence or absence of trees and the type of LiDAR source (e.g., ALS, TLS). Throughout our experiments, we vary $K$ between one and three for each point cloud, adapting to the specifics of the surveyed ecosystem. This grouping strategy naturally segments the point clouds into components such as forest ground, tree trunks/branches, or foliage, with ground having the lowest complexity and foliage exhibiting the highest, all without the need for explicit segmentation approaches.

% Pair-wise structural similarity between 
Inter-scan matching involves establishing correspondences between structural complexity levels across pairs of scans. This process is formulated as an assignment problem, ensuring a one-to-one correspondence between groups of points in the source scan and corresponding groups in the target scan. This is formulated as the assignment that minimizes the sum of pair-wise SC distances: 
\begin{equation} \label{matching}
    \hat{\A} = \argmin \sum_{k} \sum_{k'} A_{k,k'} ~ \| \text{SC}_s(k) - \text{SC}_t(k') \|_{\ell_2} 
\end{equation}
subject to the constraints that $A_{k,k'} \in \{0,1\} $ and $\sum_{k} A_{k,k'} = 1, \sum_{k'} A_{k,k'} = 1, \forall k,k'$. In the case when $K=3$ for example, $k,k' \in \{1,2,3\}$. Eq.\eqref{matching} thus finds the optimal correspondence between point-cloud groups of points, meaning it finds corresponding groups of points of similar 3D structure between the source and target point-clouds. Intuitively, we anticipate associating components such as ground-to-ground, tree trunks-to-tree trunks, and foliage-to-foliage between the source and target point clouds. 

The assignment problem, outlined in Eq.\eqref{matching}, has previously been formulated for 3D LiDAR point cloud flows \cite{castorena2020motion}. In our context, we adapt this formulation to match structural similarity levels between point clouds and solve it by leveraging existing solutions from \cite{bertsekas1988auction}.
For compactness, we represent the estimate $\hat{\A}$ through the matched indices function between source levels $k\in \{1,..,K\}$ and target levels $k' \in \{1,..,K\}$ as $\sigma: \{1,..,K\} \rightarrow \{1,..,K\}$.  %where the support of $\sigma : \Zspace \rightarrow \Zspace$ is denoted as $K=|\sigma|$. 
The function $\sigma$, thus associates a group of 3D points at a level of structural complexity in the source to a group of 3D points at the corresponding level of structural complexity in the target point cloud. %One additional note is that the structural similarities in Eq.\eqref{matching}, are computed through point range distributions with the point-to-plane projection that best fits the grouped points in each scan.

\subsubsection{Incremental Alignment Optimization}
% Incremental registration 
The proposed co-registration approach utilizes an incremental optimization strategy to iteratively refine alignment estimates. Grouped 3D points, as detailed in Section \ref{Ssec:grouping}, are co-registered in matched groups using this incremental strategy. Alignment begins with levels of lower structural complexity (e.g., $k=1$) and progresses to more complex levels (e.g., $K=3$). At each matched structural level, ICP is used as optimization on the 3D points $\X_s^{\{\Y_s==k\}}$ and the matched $X_t^{\{\Y_t==\sigma(k)\}}$. Here $\Y_s==k$ indexes only those 3D points pertaining to group $k$.
The resulting transformation $\{\R^{k}, \t^{k}\}$ is subsequently used as initialization at the next complexity level, where the 3D points associated with the next structural complexity level are aggregated to the points that have been already used in previous steps.
This incremental process continues iteratively until all structural levels up to $K$ are aligned (e.g., $K$=3). 
One important aspect to note is that at each structural level, the indexed point cloud is down-sampled (denoted as $\downarrow_{\delta}$ next to a scan $\X$) using a uniform grid with a cell size of $\delta$. This downsampling serves two primary purposes: first, it mitigates differences in point densities between scans, thereby reducing registration biases caused by variations in sampling densities (e.g., near-surface in TLS or canopy in ALS). Second, downsampling reduces the computational demands involved in computing normals, estimating parameters of the mixture of vMF distributions, and running ICP on densely sampled scan pairs. Throughout our experiments, we use a downsampling factor of $\delta_0 = 0.05$ m, but this can be adjusted based on the specific scanning density characteristics of the sensor.
In the final step, ICP is re-executed on the entire point cloud using the last computed estimates  $\{\R^K, \t^K\}$. This re-execution typically involves a few iterations (e.g., 50) with a finer downsampling resolution, such as $\delta = 0.025$ m.
A summary of the whole described co-registration approach is included in Algorithm \ref{multilevel_algo}. %Groups are formed by similarity between range point-to-plane projection distributions matched between scans acording to similarity. Matched levels are later used to incrementally register the full-scans.

% Algorithm
\begin{algorithm}[t]
	\small
	\caption{ {\small Automatic Forest structure-based alignment (ForestAlign) } } \label{multilevel_algo}
	\begin{algorithmic}[1]
		\BState \textbf{Input:} Source $ \X_s$ and target $\X_t$ point clouds.
		\State \textbf{Set:} downsampling factors $\delta_0$ = 0.05(m),  $\delta$ = 0.025(m), $r=0.25$(m) spherical neighborhood radius, structural complexity levels $K=3$ for TLS, $K=2$ for ALS, $K=1$ when no trees.
		\State \qquad  Initializations $\R^0=\mathbf{I}_{3\times 3}$ (identity),$\t^0 = \mathbf{0}_{3\times 1}$ (zero vector).
		\State \textbf{Grouping 3D points into structural complexity levels: }  
		\State \quad \textbf{for} $q$ in $\{s,t\}$ \textbf{do}:
		\State \qquad Downsample point clouds $\X'_q = \X_q{\downarrow}_{\delta_0}$
		\State \qquad 3D plane approximation $\tX_q=\text{extract-planes}(\X'_q, r)$ 
		\State \qquad // Group 3D points by structural complexity through mixture of von Mises-Fisher distribution.
		\State \qquad $\Y, \pmb{\mu}, \pmb{\kappa} = \text{mix-vMF}( \tX_q,K ) \quad \text{with} \quad \Y \in \{0,1,...,K\}^N$ labels, mean and concentration params. $\pmb{\mu}, \pmb{\kappa} \in \Rspace^K$. 
		\State \quad \textbf{end} 
		\State \quad \textbf{Output:} Structural complexity groupings $\{ \Y_s, \Y_t \}$
		\BState \textbf{Matching structural levels between scans:}
		\BState \quad Compute $\hat{\A} = \argmin \sum_{k} \sum_{k'} A_{k,k'} ~ \| \text{SC}_s(k) - \text{SC}_t(k') \|_{\ell_2}$ (Eq. \eqref{matching}) \quad and obtain $\sigma$.
		\BState \textbf{Incremental co-registration:}   
		\BState \quad \textbf{for} $k$ in $[1,...,K]$ \textbf{do}: 
		\State \qquad $\{\R^{k},  \t^{k} \} = \mathbf{ICP} ( \X'^{\{\Y_s == k\}}_s, \X'^{\{\Y_t == \sigma(k)\}}_t, \{\R^{k-1},  \t^{k-1} \} )$
		\State \quad \textbf{end}
		\State \quad \textbf{run:} $\{\R,  \t \} = \mathbf{ICP} ( \X_s{\downarrow}_{\delta}, \X_t{\downarrow}_{\delta}, \{\R^{K},  \t^{K} \} )$
		\State \textbf{Output: } Rotation and translation $\{\R, \t\}$.
	\end{algorithmic}
\end{algorithm}

\section{Experiments}
\label{Sec:experimentation} 

% Data, equipment, environment and summary
In the following set of experiments, validation is conducted on real LiDAR point cloud datasets collected from various forest ecosystems. Specifically, the experiments include co-registration of: (1) multi-view TLS scans, and (2) co-located TLS and ALS scans. The datasets were primarily sourced from the Bandelier National Monument in New Mexico, USA, encompassing high-elevation mixed-conifer forests, and from Ft. Stewart, GA, USA, unless otherwise noted. The implementations were carried out in Python, utilizing the VTK-based PyVista module for visualization, the scikit-learn implementation of the mixture of vMF algorithms \cite{banerjee2005clustering}, and selected tools from Open3D \cite{Zhou2018}, including ICP and downsampling functionalities.

\subsection{Datasets}
\label{SSec:dataset}
%
%\begin{figure} [ht]
%	\centering 
%	\subfloat[Nearly flat terrain ]{\includegraphics[width=0.2\linewidth]{density_of_points}}
%	\subfloat[Steep terrain ]{\includegraphics[width=0.2\linewidth]{density_of_points_steep}}
%	\caption{BLK360 LiDAR point cloud distribution.}
%	\label{fig:distribution} 
%\end{figure}
%
% TLS dataset
The TLS datasets were collected in between 2020-2024 using a Leica BLK360 LiDAR with a laser scanner operating at an 830 nm wavelength and fields of view (FOV) of 300$^{\circ}$ in the vertical and 360$^{\circ}$ in the horizontal directions. The LiDAR was mounted on a static tripod during scanning. The center locations of forest inventory plots (FIP) served as reference points for positioning the "center" LiDAR scan, with multi-view scans placed within the FIP quadrants defining the total FIP area coverage. Each inventory plot measured 50 x 20 meters, with LiDAR scans positioned at the center and distributed approximately 15 x 5 meters apart from the center in each of the four plot corners. Each scan captured approximately 8 million points, with around 4.6 million points reflected from the ground. It was observed that the majority of returns were concentrated below the 15-meter range overall, with most returns specifically below 10 meters for the ground surface.

% ALS dataset
%The ALS data was collected in October 2021 by TetraTech over 170 square miles around Los Alamos, NM. 
The ALS system featured a Galaxy T2000 LiDAR sensor mounted on a fixed-wing aircraft. This LiDAR operates with a 1064 nm wavelength laser and can capture up to 8 multiple returns per laser pulse. Its scan field of view (FOV) ranges from 10-60$^{\circ}$, with an accuracy ranging from less than 0.03 to 0.25 meters RMSE within the 150 to 6500 meter range. The surveyed area was covered using multiple flight lines, including 7 cross strips to achieve a point density of 10 to 15 points per square meter. It's worth noting that the ALS datasets utilized in this study include all 8 multiple returns per laser pulse, which enhances point return density especially under canopy cover. Unlike TLS, ALS scans are not affected by satellite signal occlusion and were aligned using a Trimble CenterPoint RTX RTK positioning system, providing centimeter-level accuracy.

% RMSE metric
{\it Performance Metrics:} All quantitative evaluations in our study were conducted using the registration root mean square error (RMSE) metric. The registration RMSE was measured relative to the ground truth six degrees of freedom orientation and position registration parameters. 
Ground truth was obtained, whenever additional source data was available, by cross-validation.

In cases of overlapping TLS-to-TLS pair-wise co-registration, the ground truth parameters were obtained by first manually co-registering each TLS scan in the pair to the overlapping ALS data (note that, for computational efficiency, only a truncated ALS point cloud was used, not the entire regional ALS). After deriving the relative TLS-to-ALS co-registration for each scan in the pair, we calculated the relative position between the TLS scans in the ALS coordinate system by applying the TLS-to-ALS transformations. We then transformed this relative position back into the target TLS reference coordinate system using the inverse of the TLS-to-ALS transformation for the target scan. 
%Finally, we refined these estimates using ICP for 50 iterations at the fine resolution of 0.01 m on the pair of TLS scans.

In the TLS (source)-to-ALS (target) co-registration case, a similar process was followed to obtain ground truth. Here, pairs of overlapping TLS-to-TLS and corresponding truncated ALS point clouds are used. The truncated ALS removes 3D points outside a 50 × 50 meter area in the lat/lon direction, centered at a TLS GPS reference location. First, manual co-registration between TLS (source)-to-TLS (cross) was performed, and then manual co-registration between TLS (cross)-to-ALS was obtained. This process provides a transformation from the source TLS scan being co-registered to the ALS scan, on which we also apply ICP using the same settings as in the TLS-to-TLS case. 
When no additional overlap between TLS-to-TLS or TLS-to-ALS is available (as in the Ft. Stewart dataset and reference target datasets), cross-validation is not used. Instead, we proceeded directly to manual co-registration between the involved TLS (source)-to-ALS (target) or TLS (source)-to-TLS (target) pairs. %, followed by ICP refinement to obtain ground truth.

The manual co-registration method we used is from the open-source CloudCompare software \cite{girardeau2015cloudcompare}, which provides interactive visualizations to manually select corresponding points between point clouds. Given the corresponding points, the software automatically co-registers the pairs of point clouds. 
%The algorithm used to obtain the registration parameters solves the classical orthogonal Procrustes problem \cite{schonemann1966generalized}. 
Our methodology involved selecting 5 corresponding points in the source and target scans, distributed across the entire point cloud. This distribution constrains the alignment more effectively than selecting points that are closely clustered together. It is important to note that 5 corresponding points exceed the minimum requirement of 3 for the Procrustes algorithm \cite{schonemann1966generalized} to perform co-registration.
Additionally, after manual co-registration, we apply ICP with a downsampling factor of 0.015 m and a closest point threshold of 0.25 m over 50 iterations to further refine the ground truth estimates from the manual alignment.

%In Subsections \ref{Ssec:overall} to \ref{Ssec:overlap}, ground truth was established by running our approach on TLS ground scans for 10 trials per pairwise scan, averaging registration parameters, and ensuring consistency with augmented ALS. Consistency with ALS was confirmed by comparing relative orientations and positions derived from TLS scan centers projected onto the ground plane in ALS scans. It is noteworthy that ALS aligned with RTK provides accuracies at the centimeter level. Following these steps, the resulting registrations were validated by a trained expert. 

In Section \ref{Ssec:benchmark}, ground truth was obtained from a reference-based method using a LiDAR dataset collected in a forest ecosystem. Evaluations were also benchmarked against other well-known standard co-registration methods. Overlap was quantified as the percentage of points in the source scan having corresponding points in the target scan within a distance of 25 cm, relative to the total number of points in the pair of scans. For registration involving multiple scans, overlap was similarly calculated for every pairwise combination within a plot. Finally, inlier RMSE measures the minimum RMSE distance between 3D points in aligned source and target point clouds that are at least 25 cm apart. This metric provides a quantitative assessment of the proximity between aligned source and target point clouds.

% TLS-to-TLS registration examples
%
\begin{figure} [ht]
	\centering 
	%
	%\subfloat[Ex.1: Miss-aligned scan]{\includegraphics[width=0.243\linewidth]{unregistered_5}}
	%\hspace{0.1em}
	%
 
 	\subfloat[Ex.1: Miss-aligned.]{\includegraphics[width=0.177\linewidth]{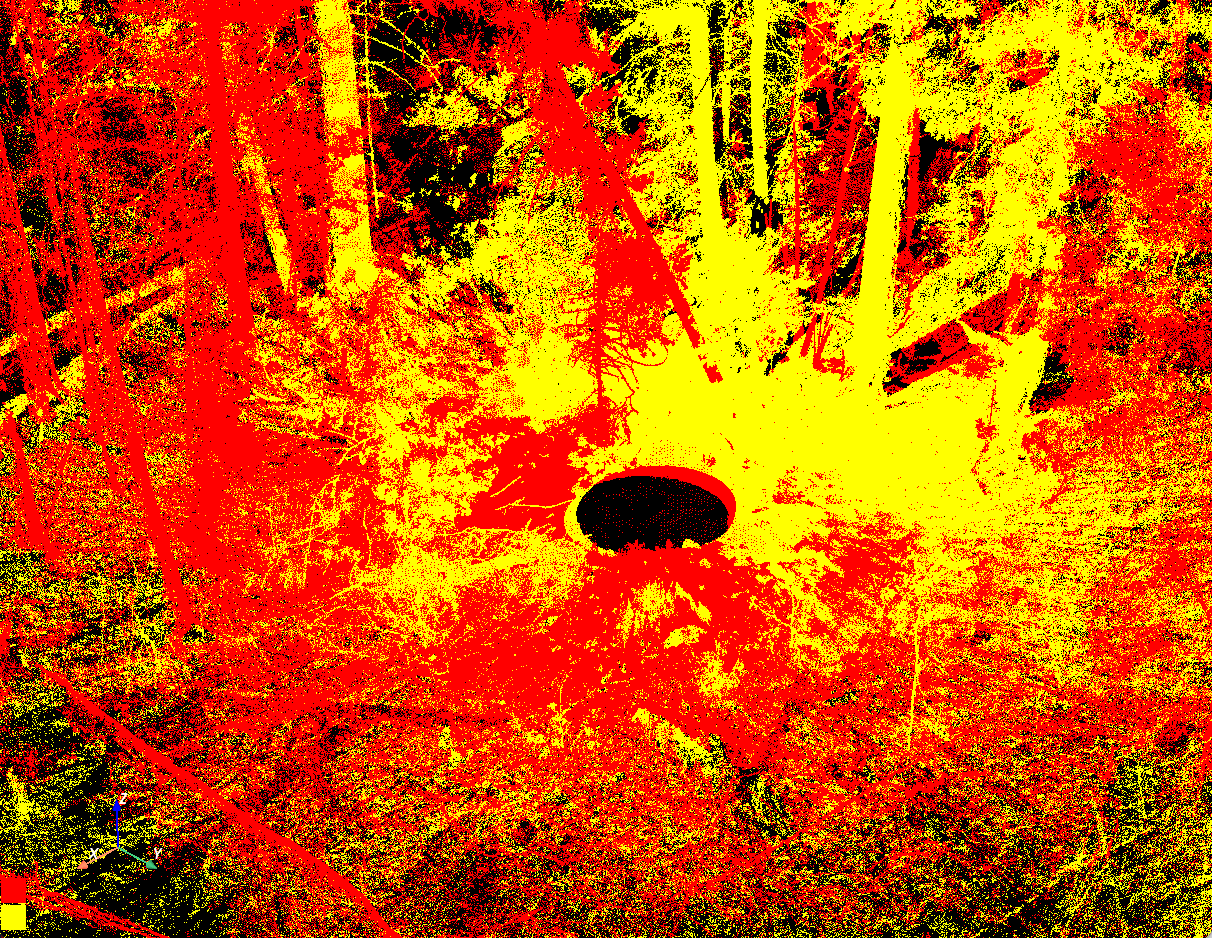}}
        \hspace{0.1em}
	%\subfloat[Ex.1: Aligned ortho/side view.]{\includegraphics[width=0.243\linewidth]{pair_ortho}}
	%\hspace{0.1em}
	%
	\subfloat[Ex.1: Aligned ortho/side view]{\includegraphics[width=0.243\linewidth]{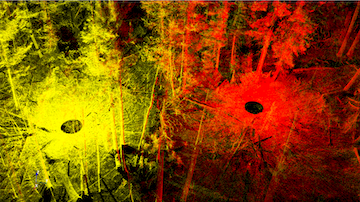}}
	\hspace{0.1em}
	\subfloat[Ex.1: Aligned ortho/side view]{\includegraphics[width=0.243\linewidth]{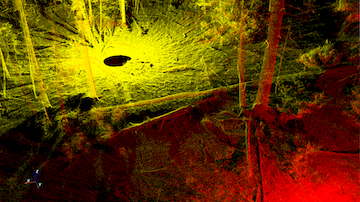}}
	\hspace{0.1em}
	\subfloat[Ex.1: Aligned side view]{\includegraphics[width=0.243\linewidth]{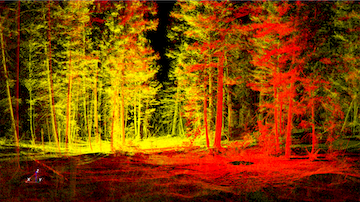}}

	\subfloat[Ex.2: Aligned ortho 5 scans]{\includegraphics[width=0.23\linewidth]{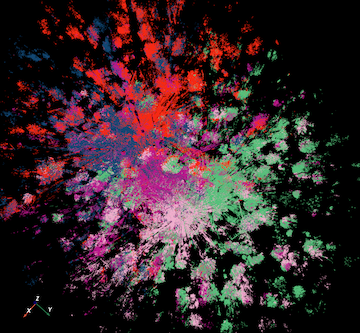}}
	\hspace{0.1em}
	\subfloat[Ex.2:  Aligned ortho 5 scans]{\includegraphics[width=0.23\linewidth]{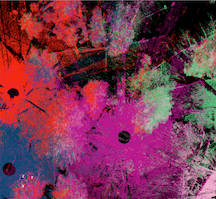}}
	%
	%\subfloat[Side-view]{\includegraphics[width=0.10\linewidth]{figures/multiview_side}}
	%
	\hspace{0.1em}
	\subfloat[Ex.2: Aligned Side-view 5 scans]{\includegraphics[width=0.452\linewidth]{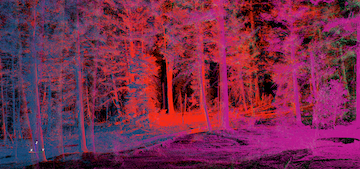}}	
	\caption{Pairwise alignment views of TLS-to-TLS co-registration. First row is an example result of pair-wise co-registration. (a) illustrates two scans miss-aligned, each in its own sensor coordinate system. (b-d) shows multiple views of the aligned point-clouds after applying our co-registration approach. Second row shows 5 aligned point clouds (using our approach), color coded each, distinctly.}
	\label{fig:pairwise_align} 
\end{figure}
%
%
%\begin{figure} [ht]
%	\centering 
%	\includegraphics[width=0.25\linewidth]{figures/pairwise_trees}
%	%
%	\caption{Pair-wise alignment at some pine tree examples.}
%	\label{fig:pairwise_align_trees} 
%\end{figure}

\subsection{ Pair-wise TLS to TLS co-registration}

\subsubsection{Overall results} \label{Ssec:overall}
The pairwise registration method described in Section \ref{Ssec:registration} was evaluated on pairs of multi-view TLS point clouds. Example results are illustrated in Fig. \ref{fig:pairwise_align}. The first row shows four views of the registration of two color-coded scans positioned 15 x 5 meters apart in a forest site featuring white fir, limber pine, aspen, Douglas fir, and surface bark, trunk, and bushes. Fig. \ref{fig:pairwise_align}(a) depicts an initial example where forest features are misaligned. In contrast, Figs. \ref{fig:pairwise_align}(b-d) display different views of the alignment results after applying our co-registration approach. The second row presents qualitative results demonstrating the alignment among five scans within a single plot using our pairwise co-registration method relative to the center point-cloud scan. Overall, Fig. \ref{fig:pairwise_align} shows well-aligned point clouds with consistent object alignment across scan transitions. Notably, we utilized three structural complexity levels ($K=3$) in the mixture of vMF distributions and empirically observed that the resulting groups mainly corresponded to: (1) the ground surface, (2) tree trunks/branches, and (3) foliage. This finding remained qualitatively consistent across various cases of forest structure captured with TLS.

%This can also be confirmed in Fig. \ref{fig:pairwise_align_trees} of a cropped and zoomed section showing trees where the trunks are mostly scanned by the red scan while leaves and branches have a consistant alignment with those from the yellow scan.

%
\begin{figure} [ht]
	\centering 
	\subfloat[Rotation roll]{\includegraphics[width=0.48\linewidth]{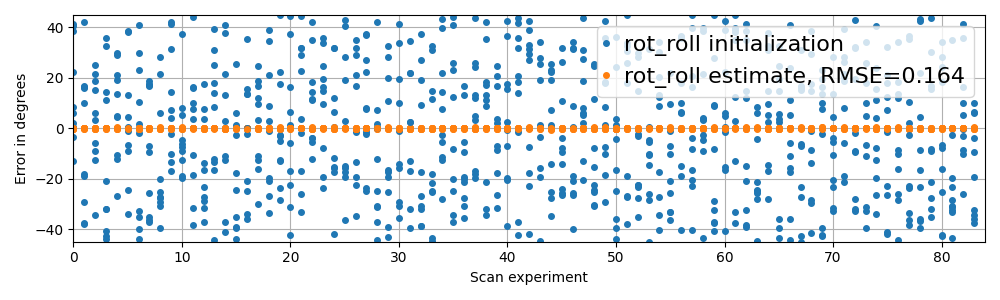}}
	\subfloat[Rotation pitch]{\includegraphics[width=0.48\linewidth]{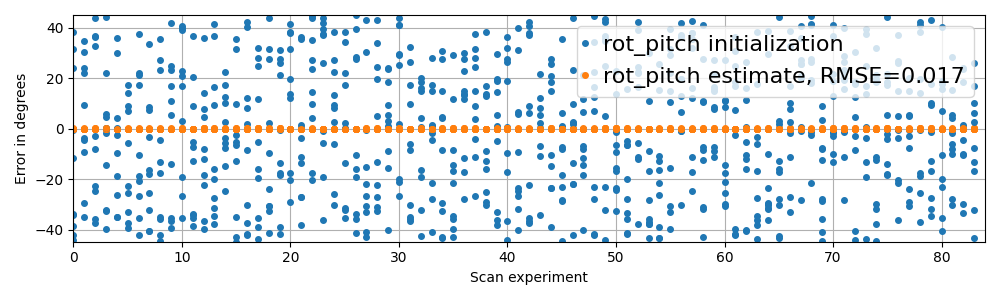}}

	\subfloat[Rotation yaw]{\includegraphics[width=0.48\linewidth]{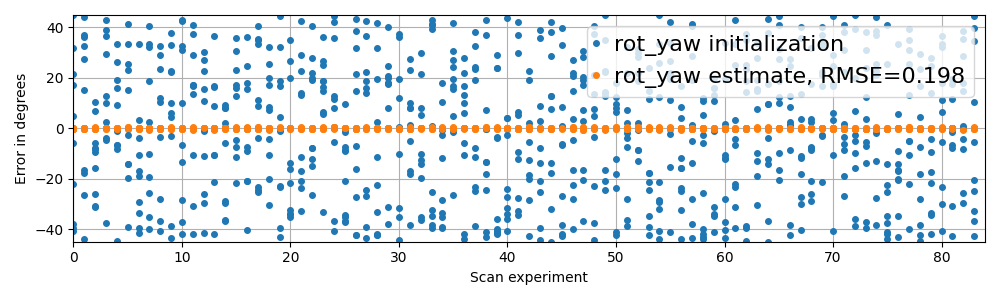}}
	\subfloat[Translation X]{\includegraphics[width=0.48\linewidth]{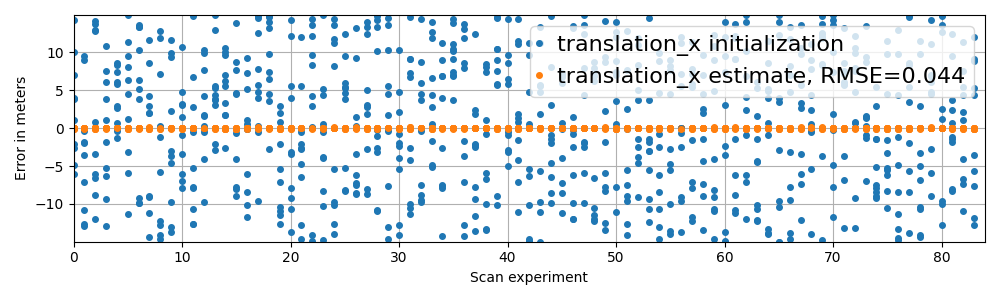}}
 
	\subfloat[Translation Y]{\includegraphics[width=0.48\linewidth]{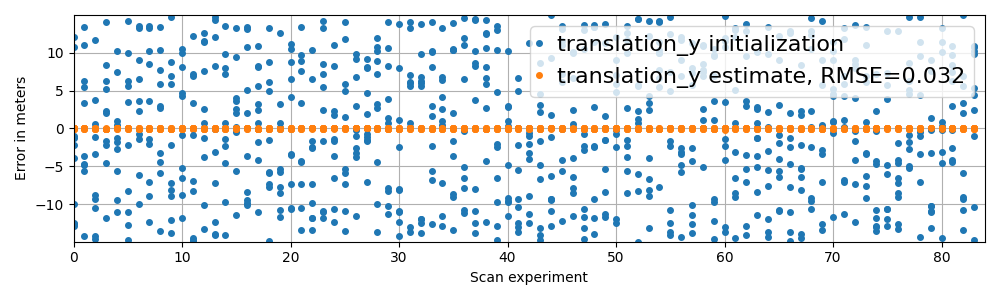}}
	\subfloat[Translation Z]{\includegraphics[width=0.48\linewidth]{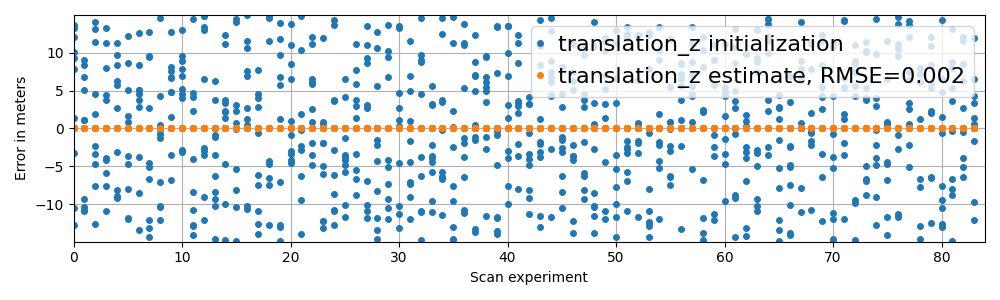}}
	\caption{Robustness of the structural complexity-based algorithm against random alignment initializations in the TLS-to-TLS co-registration problem}
	\label{fig:error_performance_exp1} 
\end{figure}
To comprehensively evaluate the performance of our proposed approach across diverse forest conditions encompassing heterogeneous vegetation and varying topographical gradients—we conducted pairwise registrations in 21 distinct inventory plots, totaling 84 pairs of scans. Initially, each scan, was synthetically misaligned by randomly and uniformly perturbing all six co-registration parameters simultaneously. These perturbations were constrained within less than $\pm 45^{\circ}$ degrees for rotations and $\pm 15$ meters for translations relative to the ground truth alignment. We performed 10 trials for each pair of scans (resulting in 840 co-registrations in total), each trial using a different initialization drawn from a uniform distribution.
%As a note, each scan pair was physically separated by a translation distance of $~$15x5 m and a randomly oriented scan axis. 
The performance of our co-registration approach across 840 co-registrations is depicted in Figure \ref{fig:error_performance_exp1}. Each point on the horizontal axis represents one trial in one of the 84 pairs of scans evaluated. The vertical axis shows the error, measured in meters or degrees, across all 10 trials for each alignment parameter (rotation and translation). The initializations are represented in blue, indicating large but random initial errors, while the final errors after co-registration are shown in orange, demonstrating successful reduction in alignment error for all six parameters.
The overall RMSE values for each co-registration parameter, computed across all scan pairs and initializations, are summarized in the second row of the legend within each plot. 
%\st{We found that the RMSE is less than $0.1^{\circ}$ for rotation and less than $0.02$(m.) meters for translation parameters.} 
We found that the RMSE is less than $0.2^{\circ}$ for rotation and less than $0.045$ meters (m) for translation parameters.
These results confirm that the method is flexible and robust, capable of handling vegetation diversity, topographical variations, and random alignment initializations effectively.
% Working on code available at: coregistration_dist.py inside the robustness folder on Darwin
%
%\begin{table}[ht]
%	\centering
%	\caption{Registration RMSE.}\label{tab:overall}
%	\begin{tabular}{lll|lll}
%		\toprule % from booktabs package
%		\multicolumn{3}{c}{ \bfseries Orientation (Degrees)}   &  \multicolumn{3}{c}{ \bfseries Position (meters)} \\
%		\bfseries \bfseries $\theta_{\text{roll}}$ & \bfseries $\theta_{\text{pitch}}$ &\bfseries $\theta_{\text{yaw}}$ &\bfseries $t_{x}$ &\bfseries $t_{y}$ &\bfseries $t_{z}$\\
%		\midrule % from booktabs package
%		0.08 & 0.04  & 0.09  & 0.014 & 0.016 & 0.0011  \\
%		\bottomrule % from booktabs package
%	\end{tabular}
%\end{table}

\subsubsection{Non-uniform ground terrain} \label{Ssec:overlap}
\begin{figure}[ht]
	\centering 
	\subfloat[Example 1: Side view of pairs of scans, physically distanced 10 meters apart, with $\approx$30$\%$ overlap.]{\includegraphics[width=0.376\linewidth]{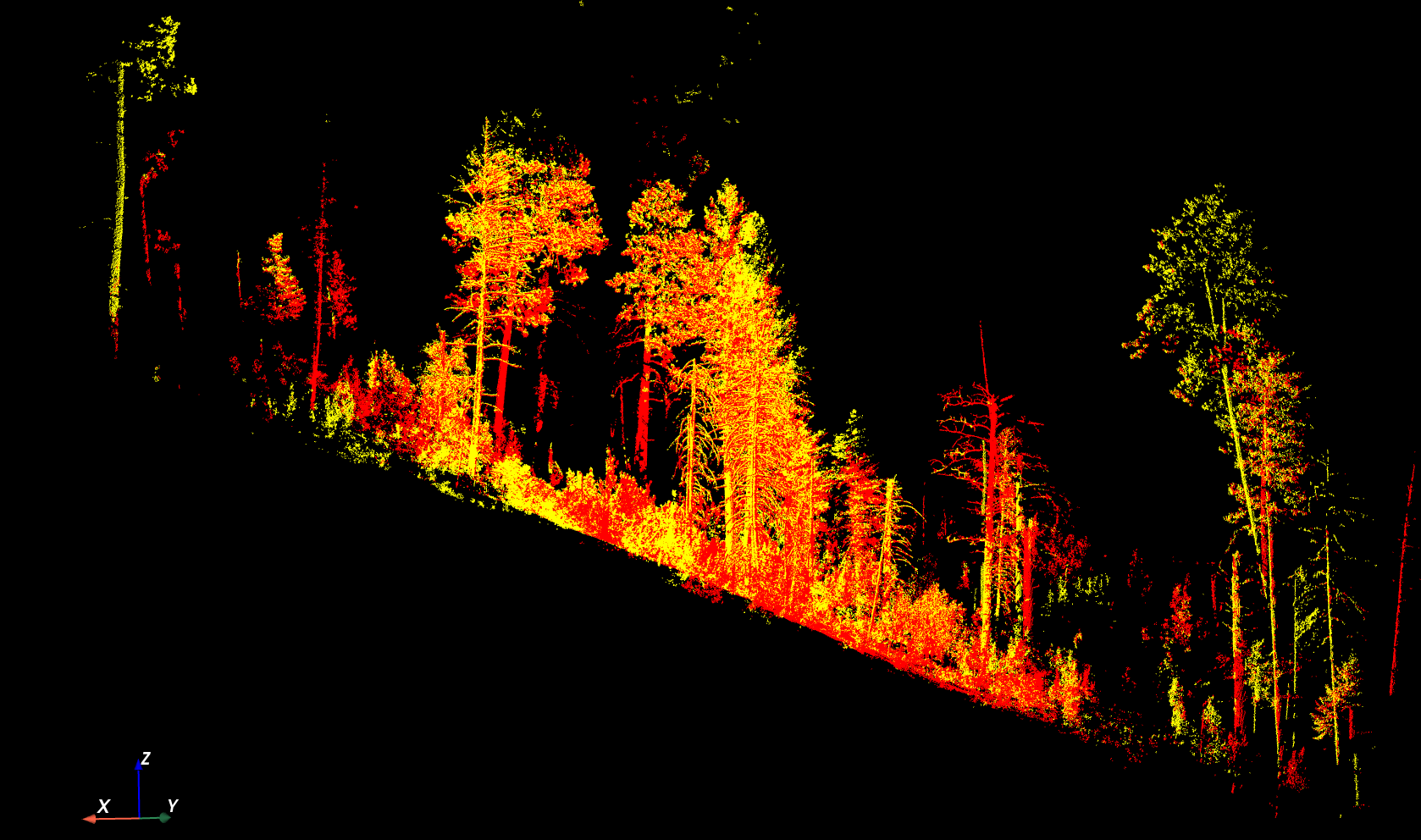}}
	\hspace{0.5em}
	\subfloat[Ex. 2: Side view of pairs of scans, distanced 15 $\times$ 5(m) apart, $\approx$15$\%$ overlap.]{\includegraphics[width=0.3\linewidth]{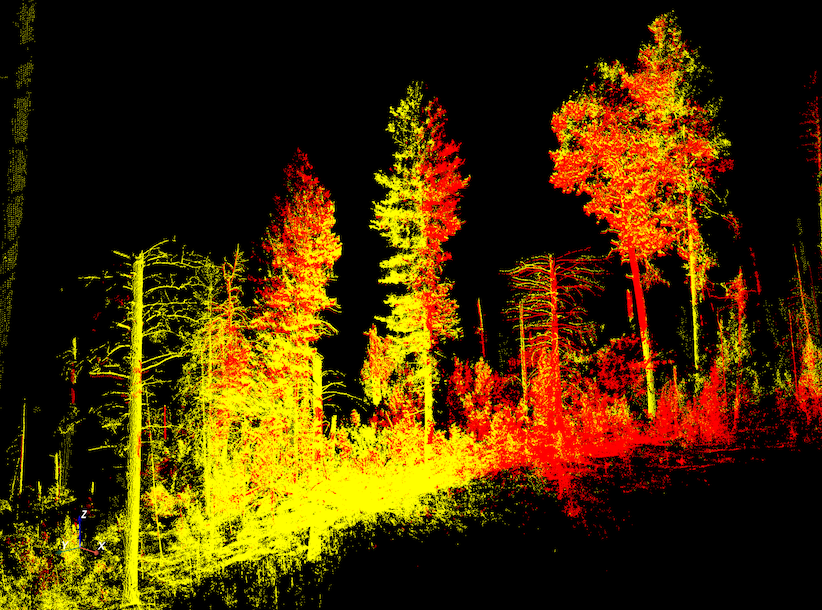}}
	\hfill
	\caption{Pairwise TLS alignment in steep terrain.}
	\label{fig:pairwise_align_steep} 
\end{figure}

In forest environments, registering TLS-to-TLS scans can be challenging, particularly in scenarios involving steep terrain. The difficulty arises not only due to the steepness of the terrain itself but also because it restricts the field of view during TLS scanning, potentially reducing the overlap between point clouds. This reduced overlap is a well-known factor that can degrade the performance of co-registration methods. To test the robustness of our structural complexity-based approach under such challenging conditions, we conducted additional experiments. Here, percent overlap is defined as the proportion of 3D points in the source point cloud that have at least a 25 cm distance to the closest point in the target point cloud, relative to the total number of points in the source when both are aligned. Fig. \ref{fig:pairwise_align_steep}.(a-b) show two representative examples of alignments produced by our method for pairs of scans 10 m apart and 15 m by 5 m apart, respectively, within a plot containing various tree species such as white fir, limber pine, aspen, Douglas fir, and ponderosa pine. Fig. \ref{fig:pairwise_align_steep}.(a) demonstrates approximately $30\%$ overlap, while Fig. \ref{fig:pairwise_align_steep}.(b) shows approximately $15\%$ overlap. Each point-cloud consists of approximately 6 million 3D points, and as such, a 1$\%$ overlap represents approximately 60000 3D points. Notably, our method consistently aligns these point clouds throughout the entire scene, as depicted in Fig. \ref{fig:pairwise_align_steep}.

\begin{figure} [ht]
	\centering 
	\subfloat[Rotation roll]{\includegraphics[width=0.32\linewidth]{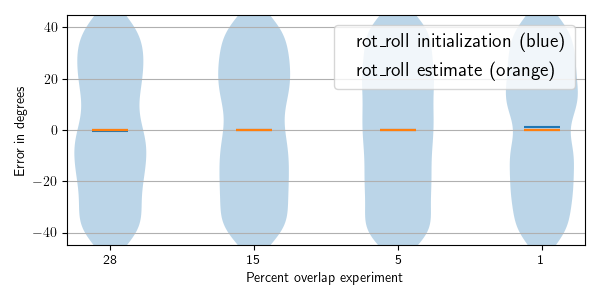}}
	\hspace{0.5mm}
	\subfloat[Rotation pitch]{\includegraphics[width=0.32\linewidth]{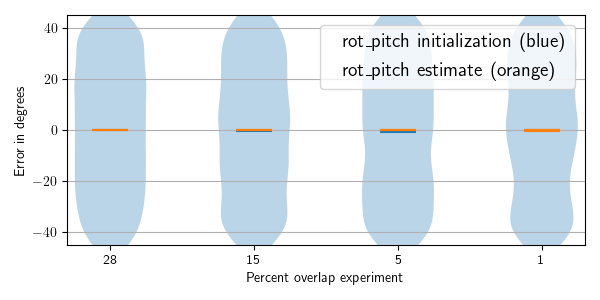}}
	\hspace{0.5mm}
	\subfloat[Rotation yaw]{\includegraphics[width=0.32\linewidth]{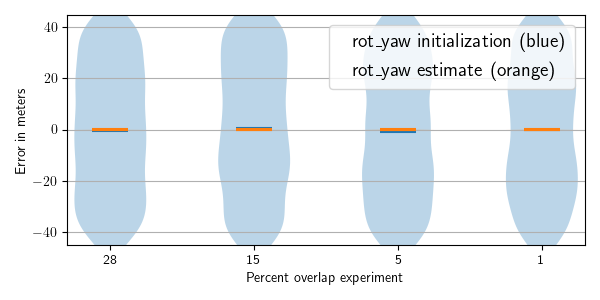}}
 
	\hspace{0.5mm}
	\subfloat[Translation X]{\includegraphics[width=0.32\linewidth]{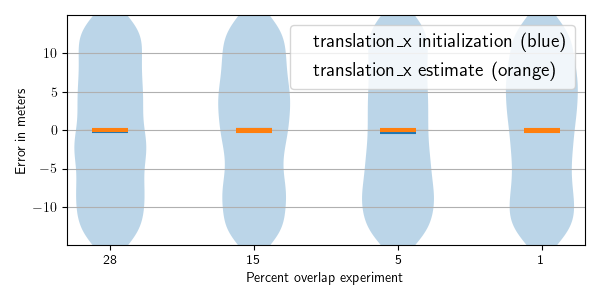}}
	\hspace{0.5mm}
	\subfloat[Translation Y]{\includegraphics[width=0.32\linewidth]{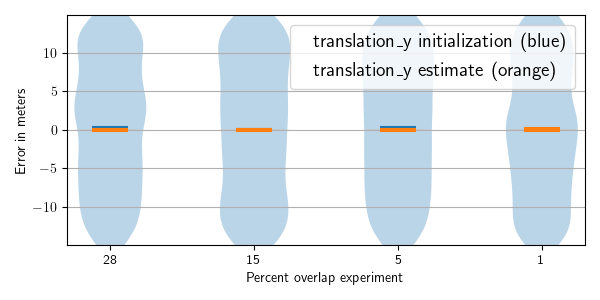}}
	\hspace{0.5mm}
	\subfloat[Translation Z]{\includegraphics[width=0.32\linewidth]{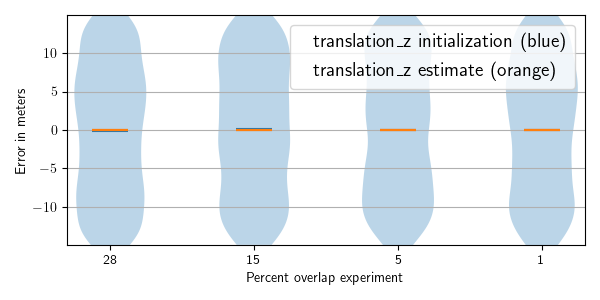}}
	\caption{Robustness of the structural complexity-based algorithm (distribution in orange) against random alignment initializations (distribution in blue) and point-cloud overlap in TLS-to-TLS co-registration.}
	\label{fig:error_performance_exp2} 
\end{figure}

Fig. \ref{fig:error_performance_exp2} provides a summary of our experiments evaluating registration performance as a function of percent overlap. Each plot corresponds to an alignment parameter, with blue and orange shapes representing the distributions of initialization and estimate errors relative to ground truth, respectively. The width of each distribution indicates the frequency distribution of error values (vertical axis), while the horizontal axis shows the percent overlap tested across four settings:  $28,15,5,1 \%$. A total of 1000 co-registration trials per percent overlap were conducted, resulting in 4000 co-registrations in total. For each trial, we randomly initialized the six alignment parameters (simultaneously randomized) from a uniform distribution centered at the true alignment, deviating by at most $\pm 45^{\circ}$ in rotation and $\pm 15$ m in translation (as shown on the vertical axis). In Fig.\ref{fig:error_performance_exp2}, it is evident that errors for all six parameters remain concentrated around zero, regardless of initialization and the percent overlap tested. Table \ref{tab:overlap} summarizes the quantitative RMSE values, providing a quantification of this error variability across different alignment parameters and percent overlaps. The consistently small RMSE values across varying levels of overlap underscore the effectiveness and robustness of our approach, even under conditions of limited point-cloud overlap.

% 
% Performance table
\begin{table}[ht]
	\centering
	\caption{Registration RMSE versus $\%$ overlap.}\label{tab:overlap}
	\begin{tabular}{r|lll|lll}
		\toprule % from booktabs package
		& \multicolumn{3}{c}{ \bfseries Orientation (Degrees)}   &  \multicolumn{3}{c}{ \bfseries Position (meters)} \\
		\bfseries Overlap & \bfseries $\theta_{\text{roll}}$ & \bfseries $\theta_{\text{pitch}}$ &\bfseries $\theta_{\text{yaw}}$ &\bfseries $t_{x}$ &\bfseries $t_{y}$ &\bfseries $t_{z}$\\
		\midrule % from booktabs package
		$\sim $28 $\%$ & 0.431 & 0.475 & 0.644 & 0.048 & 0.044 & 0.053\\
		$\sim15\%$ & 0.429 & 0.452 & 0.672 & 0.041 & 0.043 & 0.047\\
		$\sim 5\%$ & 0.561 & 0.617 & 0.713 & 0.046 & 0.041 & 0.052\\
		$\sim 1\%$ & 0.591 & 0.622 & 0.703 & 0.051 & 0.0501 & 0.064\\
		\bottomrule % from booktabs package
	\end{tabular}
\end{table}

\subsubsection{Benchmark Evaluations.} \label{Ssec:benchmark}
Here, we include a performance comparison using a benchmark dataset from \cite{gollob2020comparison}. Such dataset was collected with Styrofoam sphere targets manually and strategically placed in the scene.
This dataset consists of scans from 17 plots, each with 4 scans per plot arranged in a triangular pattern (one at the center and three at each corner 15 meters apart from the center), along with 9 manually placed targets within the scene.

Comparisons include the manual co-registration approach that uses the CloudCompare software of \cite{girardeau2015cloudcompare}, the automatic and targetless forest registration (TFR) method \cite{tremblay2018towards} , the tree-locations based method of \cite{polewski2016object}, and the canopy-based method of \cite{dai2019automated}. The TFR method from \cite{tremblay2018towards} is a trunk-based approach that identifies tree correspondences using the tree diameter at breast height (DBH) biometric and uses these to find correspondences and determine the co-registration parameters between scans. This method is commonly used as a benchmark for forest co-registration comparisons, particularly for TLS-to-TLS co-registration. 
The tree locations based method extracts the location of each tree from tree trunk information and uses relative positional information to find correspondences between scans. The canopy-based approach of \cite{dai2019automated} instead segments the tree canopies and extracts features from such segments to find correspondences between scans. In general, these methods were specifically designed for co-registering LiDAR scans from forests, relying in general on tree detection processing, and the use of these features to find correspondences between scans.
We conducted a total of 100 trials per available scan pair, initializing each trial uniformly with random rotations and translations within at most $\pm 45^{\circ}$ in rotation and $\pm 15$ m in translation error relative to ground truth. Table \ref{tab:comparison} summarizes the RMSE for all 51 pairs of scans across 100 trials each, relative to the ground truth registration parameters already provided in this case by \cite{gollob2020comparison}. Throughout all comparative performance results, those highlighted in bold represent the best performance for a given parameter, whereas the second best are highlighted in blue.
%It is important to note that TFR is representative of semantic segmentation methods and commonly used as a benchmark for comparisons.
%
% Comparison table
\begin{table}[h]
	\centering
	\caption{Registration RMSE relative to benchmark \cite{gollob2020comparison} target-based method in TLS-to-TLS}\label{tab:comparison}
	\begin{tabular}{r|lll|lll}
		\toprule % from booktabs package
		 & \multicolumn{3}{c}{ \bfseries Orientation (Degrees)}   &  \multicolumn{3}{c}{ \bfseries Position (meters)} \\
		 & \bfseries \bfseries $\theta_{\text{roll}}$ & \bfseries $\theta_{\text{pitch}}$ &\bfseries $\theta_{\text{yaw}}$ &\bfseries $t_{x}$ &\bfseries $t_{y}$ &\bfseries $t_{z}$\\
		\midrule % from booktabs package
             CloudCompare \cite{girardeau2015cloudcompare} & {\color{blue}0.0623} & \textbf{0.0462} & {\color{blue}0.041}  & \textbf{0.031} & {\color{blue}0.033} & {\color{blue}0.019} \\ 
            Trunk-based TFR \cite{tremblay2018towards} & 0.071 & 0.085  & 0.092  & 0.042 & 0.047 & 0.022\\ 
            Tree-locations \cite{polewski2016object} & 0.083 & 0.044  & 0.02  & 0.091 & 0.085 & 0.023\\ 
            Canopy-based \cite{dai2019automated}  & 0.322 & 0.621  & 0.432  & 0.153 & 0.234 & 0.191\\ 
    ForestAlign & \textbf{0.048} & {\color{blue}0.055}  & \textbf{0.032}  & {\color{blue}0.036} & \textbf{0.032} & \textbf{0.007}  \\
		\bottomrule % from booktabs package
	\end{tabular}
\end{table}
%

%\st{From Table} \ref{tab:comparison} \st{we find that our proposed framework is able to attain the accuracies of target-based methods that often yield close to perfect performance.}
From Table \ref{tab:comparison}, we see that the TFR and the tree locations method performed similarly, with both slightly better than the canopy based method. This is because both are tree trunk based, and TLS scans are known to have overall good spatial resolution below the canopy. In contrast, the canopy-based method performed decently but not as good as those that are trunk based. Main reason is that TLS is more susceptible to foliage occlusion, and such occlusion varies depending on the scan view-point. Such variation results in varying canopy estimates over scans, which makes finding feature correspondences a bit more challenging for this specific TLS-to-TLS co-registration case. Instead, the proposed approach, which leverages the entire point cloud, including the ground topography for co-registration as opposed to only tree features, performs overall better than all the tree feature based methods. It compares favorably against the manual co-registration method and achieves small error relative to the ground truth parameters provided by the benchmark dataset, and obtained using a reference based method.
%\st{This validates that our approach performs equivalently to target-based methods, while offering the advantages of full automation without the need for manually placed targets or semantic object detection.} 
%In this case, we find that our method outperforms both of the compared methods and shows favorable performance relative to the target-based method. 
In our case, however, with the additional benefits of being fully automatic, eliminating the need for manually placed targets in the scene.

\begin{figure} [ht]
	\centering 
	\subfloat[Rotation initial deviation]{\includegraphics[width=0.48\linewidth]{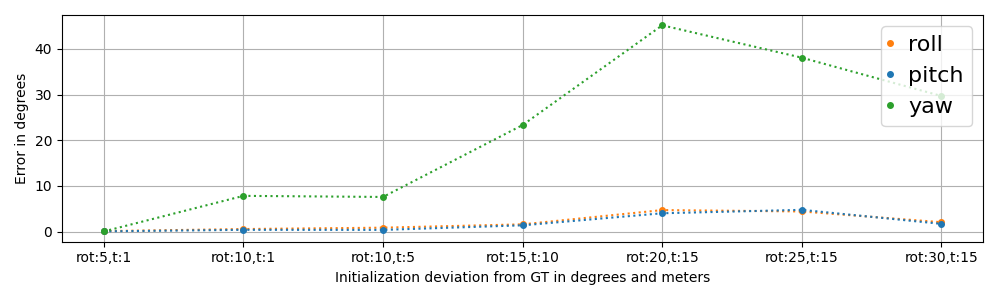}}
	\subfloat[Translation initial deviation]{\includegraphics[width=0.48\linewidth]{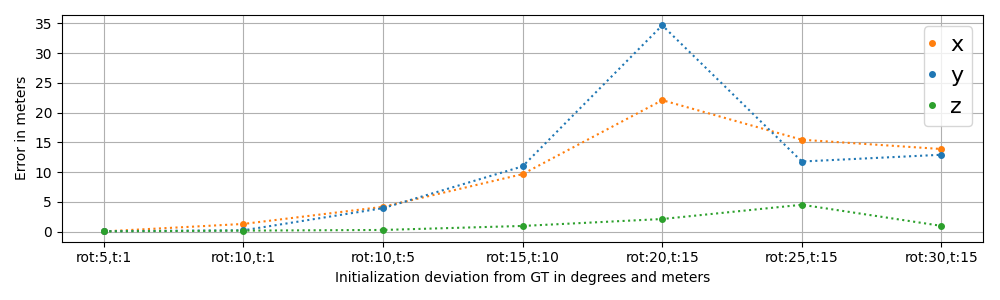}}
        \caption{Problem with standard ICP \cite{Besl:1992} for co-registering point clouds from forests as a function of deviation from ground truth.}
    \label{fig:icp_problem} 
\end{figure}

Additional experiments were conducted to compare the performance of our co-registration method against standard methods. 
The goal of adding comparisons with standard fully automated methods was to demonstrate that these methods do not perform well in forest environments.
First, we provide further analysis to show the error behavior of the most standard co-registration method (i.e., iterative closest point (ICP)) as initialization deviations increase in forest settings. While ICP is known to achieve outstanding performance in point cloud co-registration for urban scenes, this is not the case in forest point clouds, particularly in the absence of man-made components. For the range of initializations studied in this paper (e.g., within $\pm$10-15 meters in translation $\pm$45$^o$ in rotation from ground truth), ICP's performance in forest scans is less effective. 
%This range is particularly relevant for forest monitoring applications. 
To illustrate this, we conducted experiments using scans from forests in the NM dataset. We co-registered 15 scan pairs randomly selected from the dataset. Each scan was initialized according to a uniform distribution over a range of deviations from ground truth. Co-registration was performed with 50 trials per scan pair and initialization deviation. The results are summarized in Figure \ref{fig:icp_problem}.  The two plots show performance error with respect to ground truth in rotation and translation parameters on the vertical axis, as a function of initialization deviation on the horizontal axis. The plots reveal that ICP performs well within a small initialization deviation from ground truth but deteriorates as deviations increase. This issue is particularly evident for yaw rotation and $x$ and $y$ translations, where performance is more sensitive to larger deviations. ICP struggled with point clouds when translations exceeded 5 meters in translation or yaw rotations exceeded 10$^o$ from ground truth. 
In forest monitoring and surveying using TLS, standard sampling practices often place scans more than these distances apart to enhance spatial coverage and reduce occlusion. As such, standard ICP is not well-suited to the deviations typical of forest sampling protocols. The challenges with ICP in these scenarios are primarily due to the complex structure and spatial randomness of foliage, which can lead to multiple local minima in the least squares formulation of ICP, and which consequently causes ICP to get stuck at the incorrect parameters implied by the multiple local minima. Such an effect becomes more problematic at increasing levels of forest vegetation density, which in turn increase the structural complexity of the scene. 
Our approach was inspired by such ICP error behavior, and we aimed to address it by first using 3D points from low-complexity structures to obtain a rough alignment, and then incorporating points from higher-complexity structures (e.g., foliage) later, when the estimates are expected to be closer to the ground truth.

Second, we performed a more comprehensive comparison of automatic, targetless methods that do not rely on lines, contours, planes, normals, or edge features, which are typically assumed by methods designed for urban scenarios. The non-forest specific methods we compare include iterative closest point \cite{Besl:1992}, coherent point drift \cite{myronenko2010point}, gaussian mixture models \cite{jian2010robust}, support vector registration \cite{campbell2015adaptive}, trees of gaussian mixtures \cite{eckart2018fast}, filterReg \cite{gao2019filterreg}, and Bayesian coherent point drift \cite{hirose2020bayesian}, all available in the Probreg Python package \cite{probreg}. The forest specific methods we compare are the DBH trunk based TFR method from \cite{tremblay2018towards}, the tree locations from \cite{polewski2016object} and the canopy-based method from \cite{dai2019automated}.
In this case, we randomly selected 50 pairs of point clouds from the available forest plots in the NM dataset. Each pair was tested using 100 trials with random initialization from the ground truth co-registration parameters, deviating by $\pm30$ degrees and $\pm15$ meters in a uniform distribution. The RMSE relative to the six degrees of freedom was then computed and summarized in Table \ref{tab:comparison_point_methods}. The first block of rows corresponds to standard automatic co-registration methods, the second block to methods related to forest features, and the last block summarizes the performance of our approach.
% Comparison table
\begin{table}[h]
	\centering
	\caption{RMSE performance comparisons of TLS-to-TLS co-registration methods.}\label{tab:comparison_point_methods}
	\begin{tabular}{llll|lll}
		\toprule % from booktabs package
		\multirow{2}{*}{\bfseries Approach} & \multicolumn{3}{c}{ \bfseries Orientation (Degrees)}   &  \multicolumn{3}{c}{ \bfseries Position (meters)} \\
		& \bfseries \bfseries $\theta_{\text{roll}}$ & \bfseries $\theta_{\text{pitch}}$ &\bfseries $\theta_{\text{yaw}}$ &\bfseries $t_{x}$ &\bfseries $t_{y}$ &\bfseries $t_{z}$\\
		\midrule % from booktabs package
		icp \cite{Besl:1992} & 1.587 & 0.822  & 23.488  & 3.38 & 10.516 & 0.491  \\
		cpd \cite{myronenko2010point} & 0.093 & 3.81  & 22.949  & 4.865 & 10.371 & 0.654  \\
		gmm \cite{jian2010robust} & 0.153 & 0.127  & 21.921  & 4.263 & 10.611 & 0.524  \\
		svr \cite{campbell2015adaptive} & 1.81 & 3.250  & 25.481  & 5.162 & 10.731 & 0.761  \\
		gmmtree \cite{eckart2018fast} & 1.542 & 4.373  & 27.952  & 4.677 & 9.667 & 0.591  \\
		filterReg \cite{gao2019filterreg} & 0.373 & 5.982  & 29.055  & 4.362 & 10.332 & 0.823  \\
		bcpd \cite{hirose2020bayesian} & 0.328 & 4.173  & 20.940  & 4.384 & 10.010 & 2.182  \\
            \midrule % from booktabs package
            Manual co-Reg & \multirow{2}{*}{\textbf{0.21}} & \multirow{2}{*}{{\color{blue}0.27}}  & \multirow{2}{*}{{\color{blue}0.052}}  & \multirow{2}{*}{\textbf{0.033}} & \multirow{2}{*}{{\color{blue}0.036}} & \multirow{2}{*}{{\color{blue}0.002}} \\
            (CloudCompare \cite{girardeau2015cloudcompare}) &  &  & & & & \\         
            Trunk-based TFR \cite{tremblay2018towards} & 1.62 & 2.13  & 3.1  & 0.281 & 0.32 & 0.051 \\ 
            Tree locations \cite{polewski2016object} & 1.27 & 2.35  & 4.15  & 0.322 & 0.419 & 0.073 \\ 
            Canopy-based \cite{dai2019automated} & 3.59 & 4.62  & 3.1  & 0.546 & 0.419 & 0.103 \\ 
            \midrule
		ForestAlign (ours)  & {\color{blue}0.224} & \textbf{0.261}  & \textbf{0.0059}  & {\color{blue}0.0356} & \textbf{0.0349} & \textbf{0.0018}  \\
		\bottomrule % from booktabs package
	\end{tabular}
\end{table}
The second row block in Table \ref{tab:comparison_point_methods} includes results from direct manual co-registration (i.e., note that the method to obtain ground truth using cross-validation is different), as well as the results of the TFR method \cite{tremblay2018towards}, the tree locations based from \cite{polewski2016object} and the canopy-based method from \cite{dai2019automated}, the final block presents the performance of our approach.
We would like to note that in this case, significant occlusion from trees and understory debris impacted the scans due to the vegetation density of the ecosystem. 
%\sout{The TFR method \cite{tremblay2018towards} faced challenges in finding reliable correspondences between trees using the DBH metric because of such issues. However, it performed closer to the ground truth in cases with sparser understory vegetation and fewer trees.}
The TFR method \cite{tremblay2018towards} encountered challenges in finding reliable correspondences between trees using the DBH metric. It generally performed closer to the ground truth in areas with sparse understory vegetation and fewer trees but struggled in cases with dense vegetation, primarily because the prediction of tree metrics, such as DBH, becomes problematic with heavy occlusion. This is not the case of course, in forests with a sparse spatial distribution of trees as in the benchmark dataset in Table \ref{tab:comparison} where TFR performed generally well. 
The tree location method performed similarly to the TFR method. The reason is that both being trunk-based will benefit or degrade depending on the trunk segmentation and feature extraction estimation.
The canopy-based method from \cite{dai2019automated} also faced challenges with co-registration for this dataset. TLS scans taken from distances of 5-15 meters view the scene from different perspectives, and under dense forest conditions, occlusion causes significant variation in canopy feature estimates depending on scan position. In contrast, the proposed approach performed comparably to the manual co-registration method and generally exhibited small errors relative to the ground truth. We attribute this performance to the fact that our method does not rely on an intermediate feature estimation stage, which can potentially lose the rich information contained in the full point cloud. This advantage is particularly important as the accurate estimation of tree features becomes increasingly challenging and more prone to error with higher vegetation density and scene complexity.

\subsection{ Pair-wise TLS to ALS co-registration}
\begin{figure}[ht]
\centering 
    \includegraphics[width=0.90\linewidth]{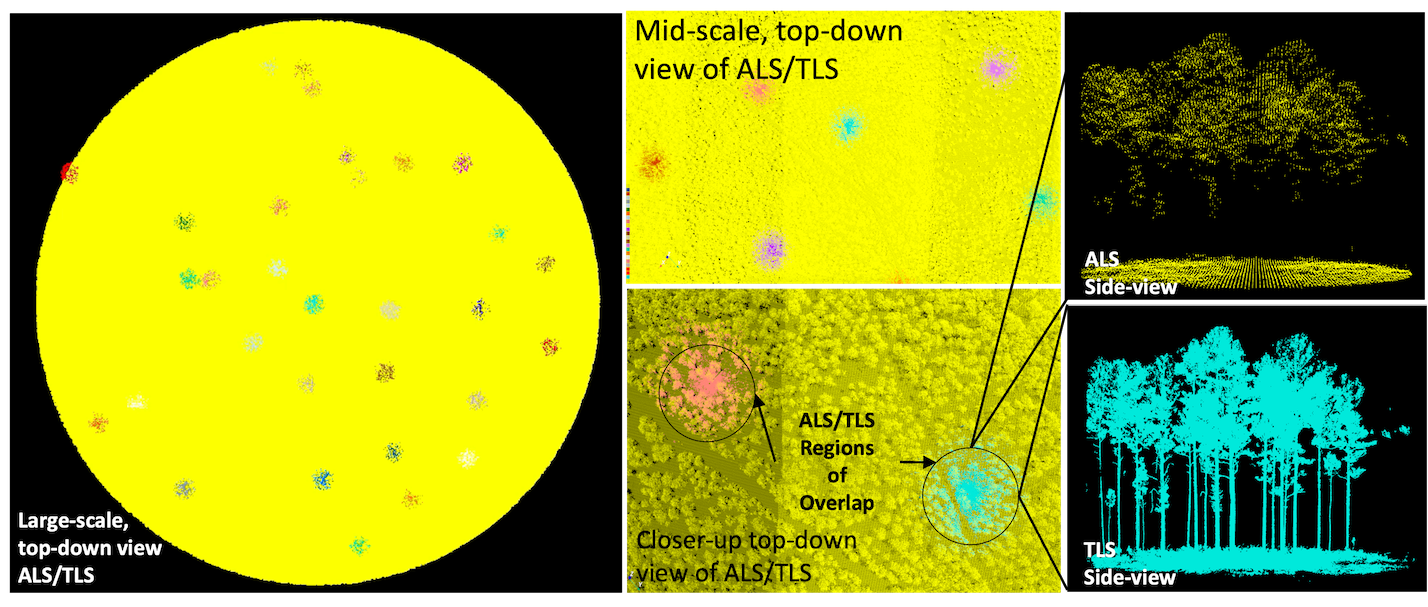}
	\caption{Co-registration of ALS (in yellow) and multiple TLS point-cloud scans at Ft. Stewart, GA forest, covering ~2.5 km in diameter with 31 TLS scans randomly distributed (a color-coded blob per scan). Overlapping ALS in top-right shows a side-view point cloud, co-registered TLS in the bottom-right shows good alignment with ALS.}
	\label{fig:tls2als_align} 
\end{figure}
Validation of our pair-wise structural complexity approach was also performed for TLS-to-ALS co-registration. Two datasets were used: the NM dataset and the Fort Stewart dataset. The NM dataset includes co-located ALS scans along with a total of 46 TLS scans covering the region. In Fort Stewart, GA, there were 42 plot-scale TLS point clouds with regional-scale ALS data. For ALS cropping, cropped scans of 50x50 m centered around the approximate GPS location of each TLS scan center were used (used solely for ALS region cropping). TLS plots were registered against ALS, with all six alignment parameters randomly initialized simultaneously from a uniform distribution centered at the GPS position of the center scan, deviating by $\pm$15m for translation and $\pm$45$^{\circ}$ for rotation. We used three levels of structural complexity $K=3$ to group the TLS scans and two levels $K=2$ for ALS.
A representative example demonstrating the regional coverage from the Fort Stewart dataset and the result of applying our method to co-register TLS scans with ALS is shown in Fig.\ref{fig:tls2als_align}. 
In the figure, the yellow-colored point cloud represents the regional ALS scan, while the differently colored blobs depict individual TLS point cloud scans with a radius of approximately 50 meters each. The middle column provides a zoomed-in view highlighting the regions of overlap between scans, while the right-most column shows a side-by-side comparison of the co-registered ALS (top) and TLS (bottom) scans at the plot scale. Despite the different spatial resolutions, ALS appearing sparser, our approach effectively aligns both point clouds, as evident from the zoomed-in comparison in the right-most column at the forest plot scale. In terms of computational performance, it took approximately 3 minutes to co-register each pair of scans, totaling 95 minutes to co-register the entire set of 31 pairs of scans in the Fort Stewart dataset.

\begin{figure} [ht]
	\centering 
	\subfloat[Ft.Stewart dataset, rotation parameter]{\includegraphics[width=0.45\linewidth]{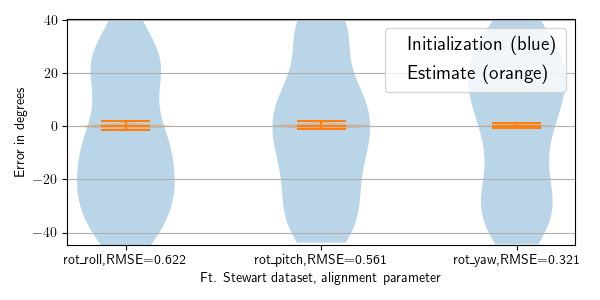}}
	\hspace{0.5mm} 
	\subfloat[NM dataset, rotation parameter]{\includegraphics[width=0.45\linewidth]{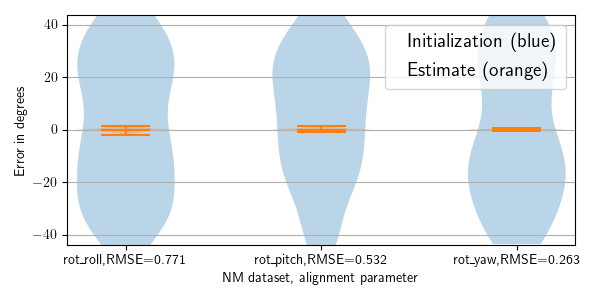}}

 	\subfloat[Ft.Stewart dataset, translation parameter]{\includegraphics[width=0.45\linewidth]{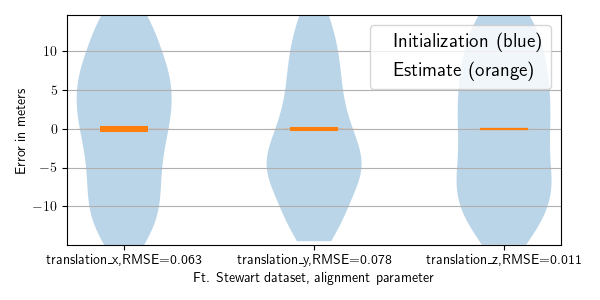}}
	\hspace{0.5mm}
	\subfloat[NM dataset, translation parameter]{\includegraphics[width=0.45\linewidth]{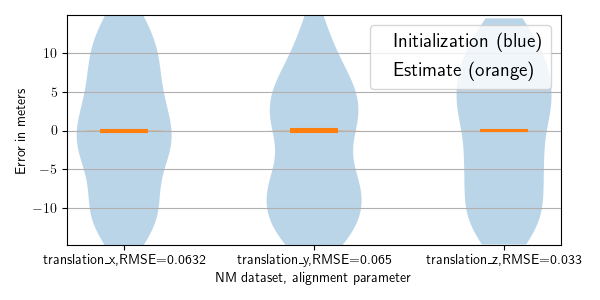}}
 
	\caption{Robustness of the structural complexity-based algorithm (distribution in orange) against random alignment initializations (distribution in blue) in TLS to ALS co-registration.}
	\label{fig:error_performance_exp3} 
\end{figure}
Quantitative evaluation of TLS-to-ALS co-registration performance was conducted using registration RMSE relative to ALS orientation as ground truth, as described in Section \ref{SSec:dataset} (Note that target-based methods for obtaining ground truth is impractical in TLS and ALS co-registration because of a 2 year time disparity between the ALS and the TLS collection). We evaluated performance in both the NM and the Ft. Stewart datasets, where for each pair we generate a total of 100 random initializations. Fig. \ref{fig:error_performance_exp3} illustrates the distribution of initializations (in blue) and the error estimates (in orange) from our proposed approach. The vertical axis represents error frequency, while the horizontal axis shows each alignment parameter out of the six. Fig.\ref{fig:error_performance_exp3}.a,c summarize the rotation and translation results for the Ft. Stewart dataset, and Fig. \ref{fig:error_performance_exp3}.b,d for the NM dataset, respectively. In all cases, the orange plots corresponding to estimation error are tightly concentrated around zero with minimal variability. RMSE values displayed in the horizontal labels for each parameter demonstrate that our approach achieves precise registration relative to ground truth. 
%\st{In both datasets, our method achieved an RMSE of less than $0.6^{\circ}$ for rotation and $0.07$m for translation.} 
In both datasets, our method achieved an RMSE of less than $0.6^{\circ}$ for rotation and $0.075$m for translation.
%Slightly better performance in the Ft. Stewart dataset can be attributed to its higher ALS spatial resolution of approximately 15 points per square meter compared to 10 points per square meter in the NM dataset, partly due to some areas being surveyed twice in the former dataset. 
It is noteworthy that registration errors in TLS-to-ALS are slighlty larger compared to pair-wise TLS-to-TLS co-registration, which we attribute to TLS's higher spatial resolution compared to ALS in general. 
For structural complexity settings, we employed $K=3$ levels for grouping TLS scans and $K=2$ for ALS. This allowed our approach to effectively match 3D points associated with lower complexity (e.g., ground) between ALS and TLS, while leaving out TLS points corresponding to tree trunks/branches, which are sparse or absent in ALS due to its challenges in returning points from the mid-story, as observed in the upper-right part of Fig. \ref{fig:tls2als_align}.

Tables \ref{tab:validationTLS2ALS_NM}-\ref{tab:validationTLS2ALS_FtStewart} present a comparison between forest-based co-registration methods and the proposed approach. The methods compared include the trunk-based TFR method \cite{tremblay2018towards}, the tree-locations method \cite{polewski2016object}, and the canopy-based method \cite{dai2019automated} across the NM and Ft. Stewart datasets.
From the results, the trunk-based TFR and tree-locations methods faced the greatest challenges. This difficulty arises primarily because ALS data, with its limited ability to penetrate the canopy, produces a sparse distribution of 3D points in the mid-story, offering limited information. Of the two methods, TFR appeared to perform slightly better, likely due to fewer challenges in estimating DBH compared to determining relative tree positions and correspondences.
In contrast, the canopy-based method and the proposed approach demonstrated more favorable co-registration error. ALS data provides uniform spatial resolution across the canopy and is less affected by occlusion at this level, facilitating canopy feature extraction and reducing co-registration error. For the TLS-to-ALS case, the proposed approach exploited on this property by matching the high structural complexity of canopy points in the TLS scan with corresponding high-complexity points in the ALS data, primarily from the tree canopy. Additionally, it leveraged ground topography information to obtain a rough initial co-registration estimate, which was subsequently refined using matched points at higher complexities, primarily in the canopy.
The superior performance of the proposed method can be attributed to this approach—utilizing both ground and canopy 3D points. Overall, the proposed method outperformed the tree-based approaches and, similar to the TLS-to-TLS case, compared favorably to manual co-registration in the NM dataset.

%
%
% Comparison table
\begin{table}[h]
    %\begin{minipage}{0.45\linewidth}
	\centering
	\caption{Registration RMSE comparison in TLS-to-ALS in NM dataset}\label{tab:validationTLS2ALS_NM}
	\begin{tabular}{r|lll|lll}
		\toprule % from booktabs package
		 & \multicolumn{3}{c}{ \bfseries Orientation (Degrees)}   &  \multicolumn{3}{c}{ \bfseries Position (meters)} \\
		 & \bfseries \bfseries $\theta_{\text{roll}}$ & \bfseries $\theta_{\text{pitch}}$ &\bfseries $\theta_{\text{yaw}}$ &\bfseries $t_{x}$ &\bfseries $t_{y}$ &\bfseries $t_{z}$\\
		\midrule % from booktabs package
             CloudCompare \cite{girardeau2015cloudcompare} & \textbf{0.231} & \textbf{0.341}  & {\color{blue}0.391}  & \textbf{0.042} & \textbf{0.041} & \textbf{0.024}\\ 
            TFR \cite{tremblay2018towards} & 1.567 & 2.374  & 3.432  & 0.612 & 0.482 & 0.674\\ 
            Tree-locations \cite{polewski2016object} & 1.865 & 2.912  & 3.914  & 0.528 & 0.51 & 0.614\\ 
            Canopy-based \cite{dai2019automated} & 1.067 & 0.91  & 0.57  & 0.268 & 0.317 & 0.395\\ 
  ForestAlign & {\color{blue}0.771} & {\color{blue}0.532}  & \textbf{0.263}  & {\color{blue}0.0632} & {\color{blue}0.065} & {\color{blue}0.033}  \\
		\bottomrule % from booktabs package
	\end{tabular}
   %\end{minipage}
\end{table}
\begin{table}[h]
	\centering
	\caption{Registration RMSE comparison in TLS-to-ALS in Ft. Stewart dataset}\label{tab:validationTLS2ALS_FtStewart}
	\begin{tabular}{r|lll|lll}
		\toprule % from booktabs package
		 & \multicolumn{3}{c}{ \bfseries Orientation (Degrees)}   &  \multicolumn{3}{c}{ \bfseries Position (meters)} \\
		 & \bfseries \bfseries $\theta_{\text{roll}}$ & \bfseries $\theta_{\text{pitch}}$ &\bfseries $\theta_{\text{yaw}}$ &\bfseries $t_{x}$ &\bfseries $t_{y}$ &\bfseries $t_{z}$\\
		\midrule % from booktabs package
              TFR \cite{tremblay2018towards} & 1.142 & 0.935  & 2.14  & 0.31 & 0.481 & 0.181\\ 
            Tree-locations \cite{polewski2016object} & 2.754 & 1.59  & 1.82  & 0.48 & 0.53 & 0.3\\ 
            Canopy-based \cite{dai2019automated} &{\color{blue} 0.981} & {\color{blue}1.03}  & {\color{blue}0.79}  & {\color{blue}0.271} & {\color{blue}0.162} & {\color{blue}0.147}\\ 
  ForestAlign & \textbf{0.622} & \textbf{0.561}  & \textbf{0.321}  & \textbf{0.063} & \textbf{0.078} & \textbf{0.011}  \\
		\bottomrule % from booktabs package
	\end{tabular}
\end{table}
%
%
%\begin{table}[ht]
    %
%    \begin{minipage}{0.5\linewidth}
%	\centering
%	\caption{TLS to ALS registration RMSE \\ \textbf{Ft. Stewart dataset}}\label{tab:tls2als_FT}
%	\begin{tabular}{lll|lll}
%		\toprule % from booktabs package
%		\multicolumn{3}{c}{ \bfseries Orientation (Degrees)}   &  \multicolumn{3}{c}{ \bfseries Position (meters)} \\
%		\bfseries \bfseries $\theta_{\text{roll}}$ & \bfseries $\theta_{\text{pitch}}$ &\bfseries $\theta_{\text{yaw}}$ &\bfseries $t_{x}$ &\bfseries $t_{y}$ &\bfseries $t_{z}$\\
%		\midrule % from booktabs package
%		0.35 & 0.32  & 0.23  & 0.03 & 0.03 & 0.02  \\
%		\bottomrule % from booktabs package
%	\end{tabular}
 %   \end{minipage}
    %
%    \begin{minipage}{0.5\linewidth}
%	\centering
%	\caption{TLS to ALS registration RMSE \\ \textbf{NM dataset}}\label{tab:tls2als_NM}
%	\begin{tabular}{lll|lll}
%		\toprule % from booktabs package
%		\multicolumn{3}{c}{ \bfseries Orientation (Degrees)}   &  \multicolumn{3}{c}{ \bfseries Position (meters)} \\
%		\bfseries \bfseries $\theta_{\text{roll}}$ & \bfseries $\theta_{\text{pitch}}$ &\bfseries $\theta_{\text{yaw}}$ &\bfseries $t_{x}$ &\bfseries $t_{y}$ &\bfseries $t_{z}$\\
%		\midrule % from booktabs package
%		0.57 & 0.63  & 0.26  & 0.06 & 0.04 & 0.03  \\
%		\bottomrule % from booktabs package
%	\end{tabular}
 %   \end{minipage}
%\end{table}
%

% Discussion
\section{Discussion}
\label{Sec:discussion}

\begin{figure} [ht]
	\centering 
	\subfloat[Complexity 1: $K=3$]{\includegraphics[width=0.19\linewidth]{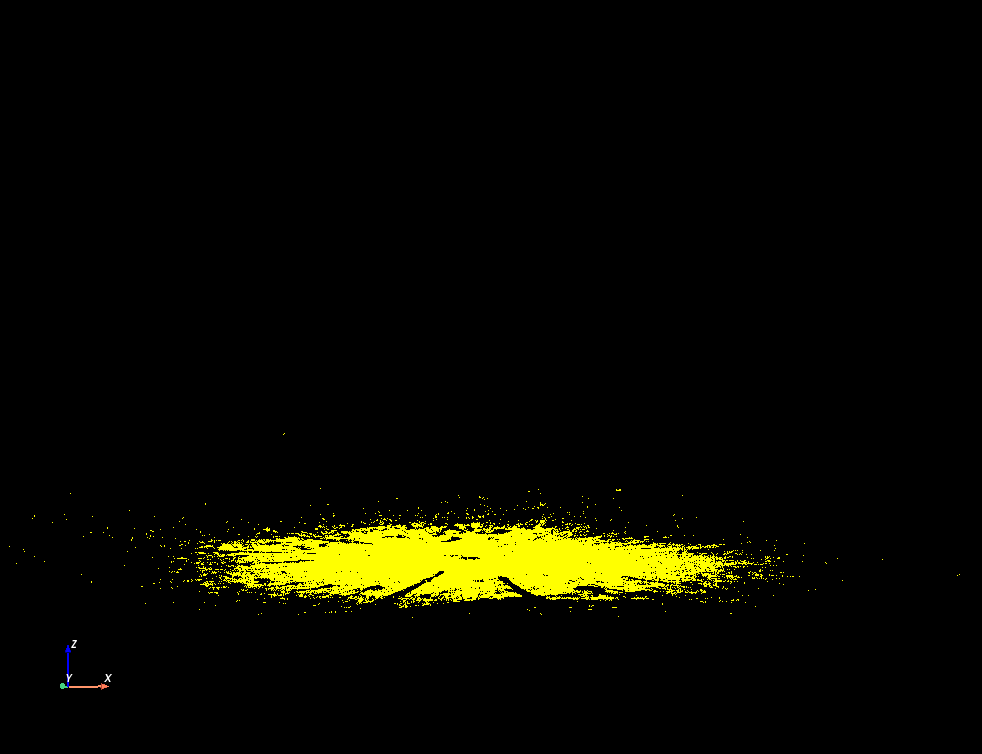}}
	\hspace{0.5mm}
 	\subfloat[Complexity 2: $K=3$]{\includegraphics[width=0.19\linewidth]{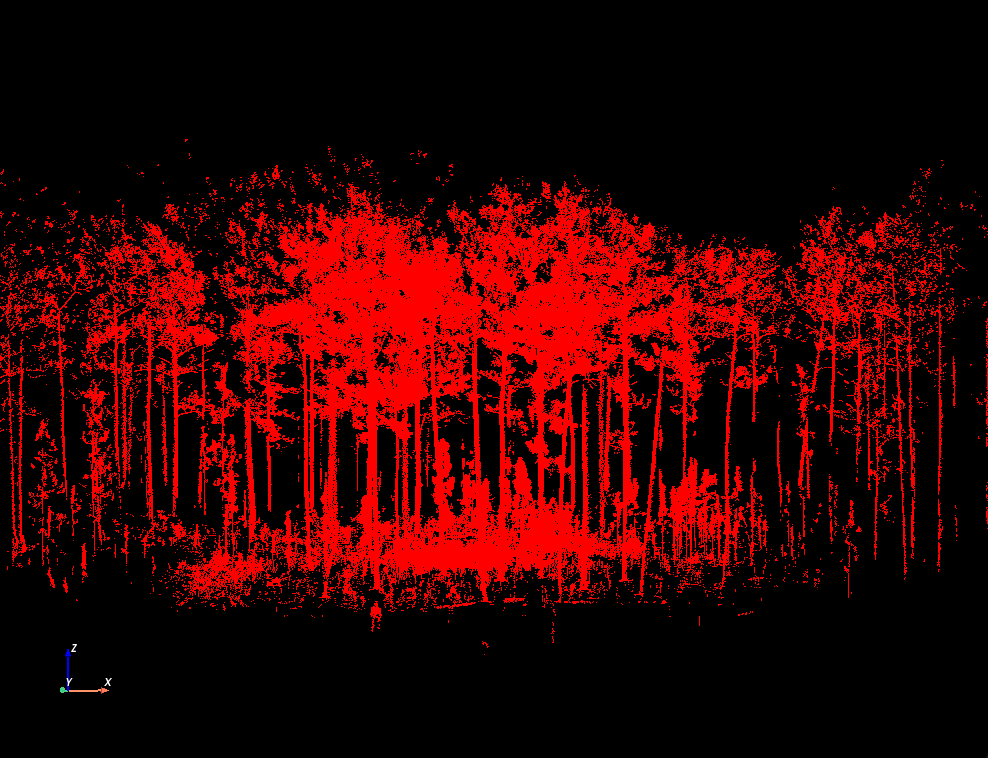}}
	\hspace{0.5mm}
	\subfloat[Complexity 3: $K=3$]{\includegraphics[width=0.19\linewidth]{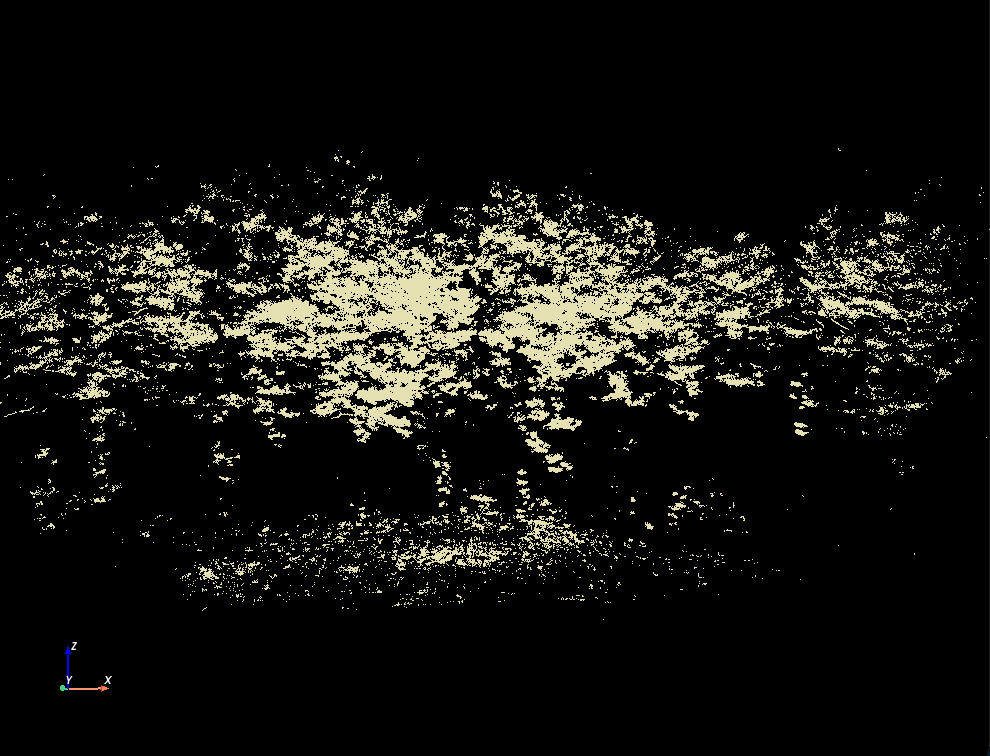}}
	\hspace{0.5mm}
	\subfloat[Complexity 1,3: $K=3$]{\includegraphics[width=0.19\linewidth]{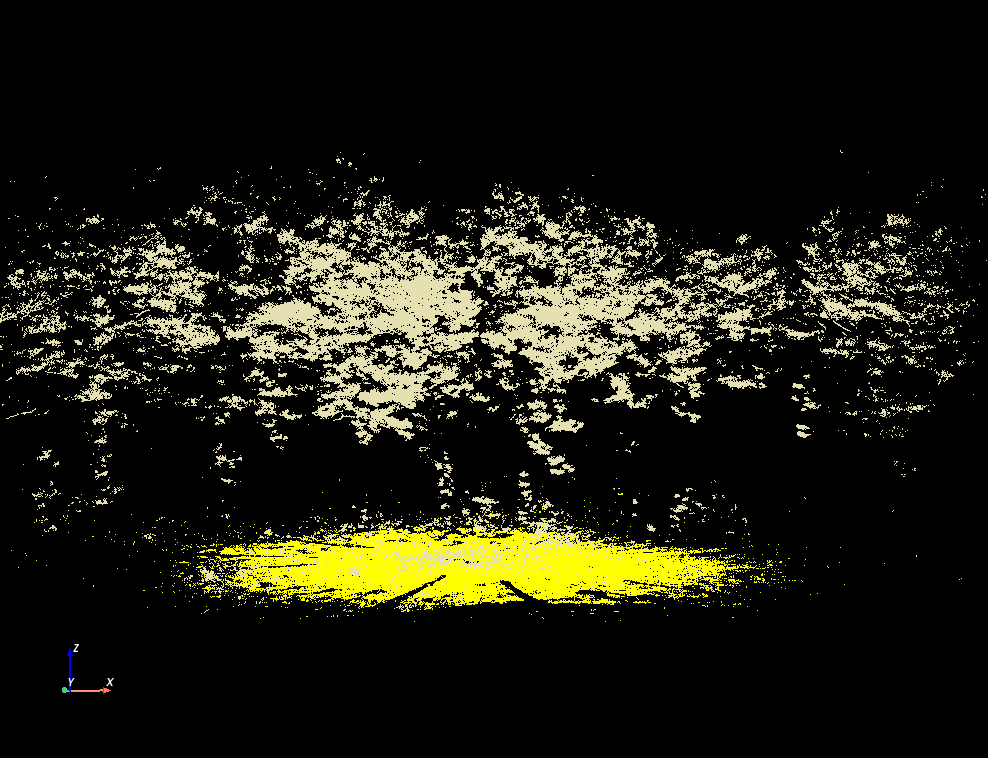}}
	\hspace{0.5mm}
	\subfloat[Complexity 1-3: $K=3$]{\includegraphics[width=0.19\linewidth]{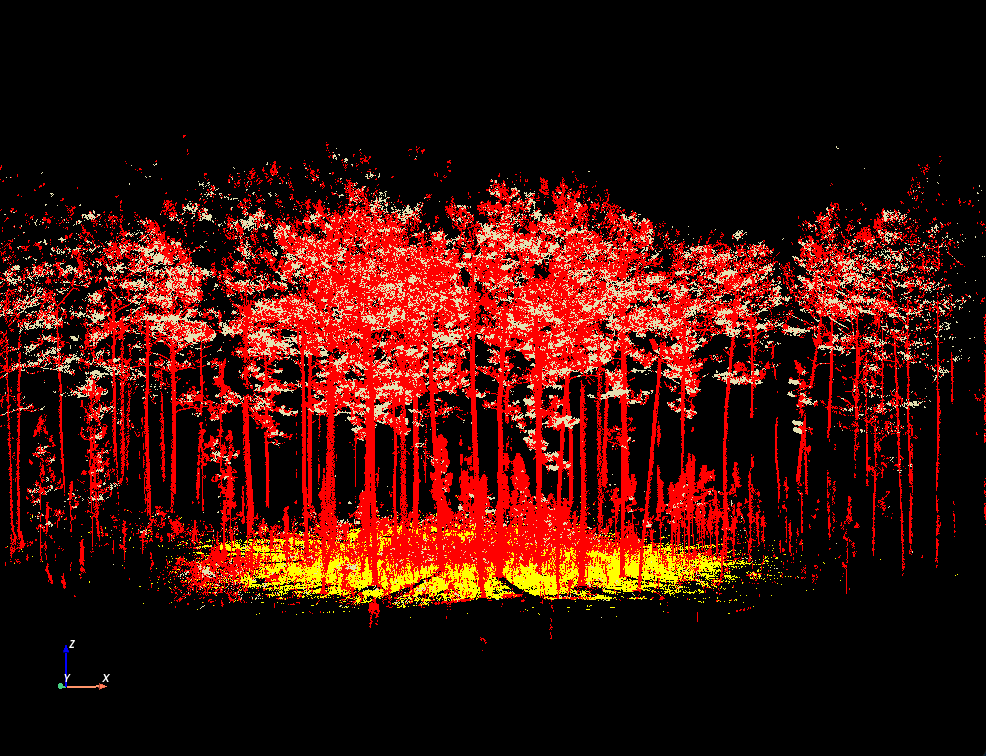}}

	\subfloat[Complexity 1: $K=2$]{\includegraphics[width=0.3\linewidth]{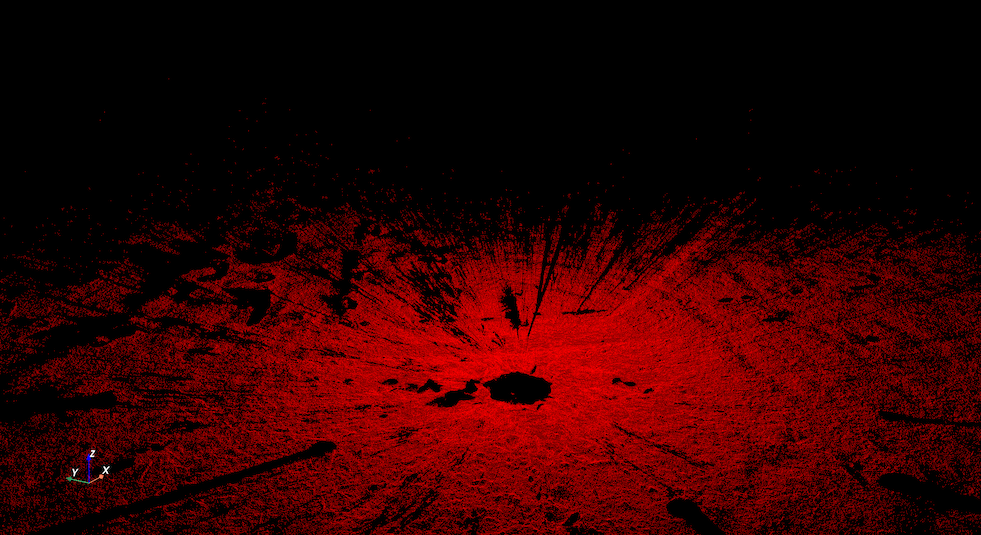}}
 \hspace{0.25mm}
	\subfloat[Complexity 2: $K=2$]{\includegraphics[width=0.3\linewidth]{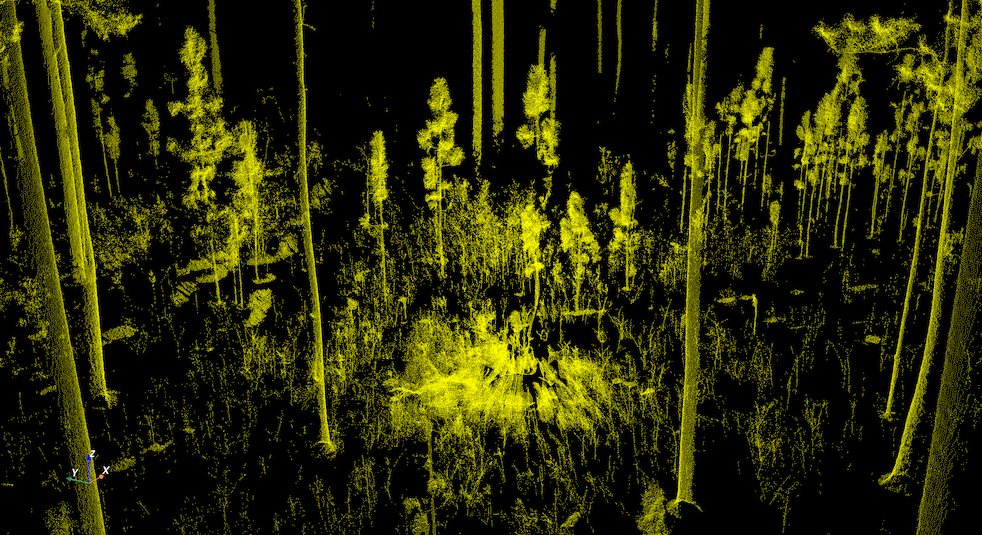}}
 \hspace{0.25mm}
	\subfloat[Complexity 1 and 2: $K=2$]{\includegraphics[width=0.3\linewidth]{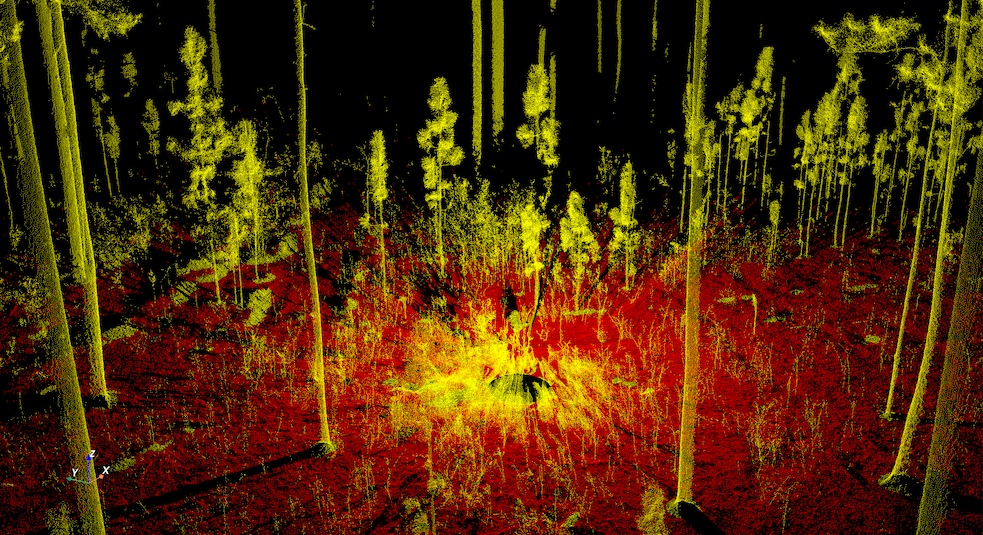}}
	\caption{Representative 3D points are grouped by structural complexity into different sets of $K$-groups. For $K=2$, our approach automatically groups 3D points into ground and above-ground vegetation. Similarly, with $K=3$, points are classified into ground, tree trunks/branches/small shrubs, and foliage effectively partitioning the underlying forest components.
        }
	\label{fig:structural_complexity} 
\end{figure}

% What is the intuition why the incremental strategy based on structural complexity?
%The intuition behind our approach lies in our assumption that a forest is composed of objects with varying structural complexities. The 3D points that belong to objects of lesser structural complexity should be exploited first for point-cloud alignment, with possible refinements through the aggregation of 3D points grouped by increasing structural complexities. This inspired us to split the point-cloud into groups and exploit those groups incrementally for co-registration. The justification for this incremental strategy is based on our empirical findings that 3D points coming from foliage structures lock the alignment into incorrect parameters when the entire point-cloud is used at once. The proposed approach leverages this finding and exploits the structure of foliage as a refinement, only after 3D points from other less complex structures have been used to estimate a rough alignment.
%
Qualitative empirical evidence shows that when we set the number of structural complexity groups to $K=3$, it tends to group 3D points mainly into ground, tree trunks/branches/shrubs, and foliage. A representative example illustrated in Fig.\ref{fig:structural_complexity}.(a-e) corroborates this. Here, 3D points color-coded in yellow correspond to the ground surface associated with the group of lowest complexity, 3D points color-coded in beige represent the foliage at the highest complexity, while those color-coded in red correspond to tree trunks/branches at mid-structural complexity. This grouping is characteristic of our approach in TLS when $K=3$. We point out in Fig.\ref{fig:structural_complexity}.(d), that points in groups of complexity 1 and 3 mainly consist of ground and foliage, similar to the point clouds captured by ALS, characterized by a sparse resolution in the mid-story as seen in the ALS side view in the top-right of Fig.\ref{fig:tls2als_align}. This finding was exploited by our approach when co-registering TLS to ALS. Specifically, our assignment problem in Eq.\eqref{matching} matched the lowest and highest complexity in TLS to those of ALS. In the case of $K=2$ shown in Fig.\ref{fig:structural_complexity}(f-h), we observed that the mixture of vMF distribution groups 3D points mainly into two groups: ground and above-ground vegetation.
%

%
%TODO: Include an illustration showing the splitting of point-clouds using k-means.
\begin{figure} [ht]
	\centering 
	\subfloat[Complexity 1: $K=3$]{\includegraphics[width=0.303\linewidth]{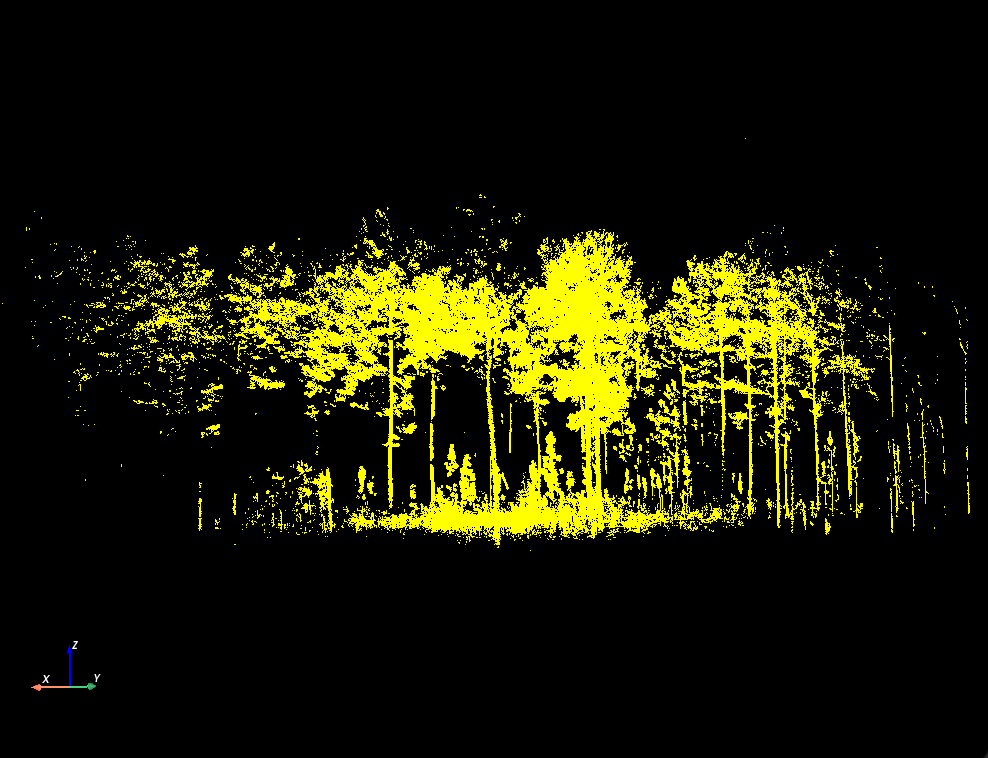}}
 \hspace{0.25mm}
	\subfloat[Complexity 2: $K=3$]{\includegraphics[width=0.304\linewidth]{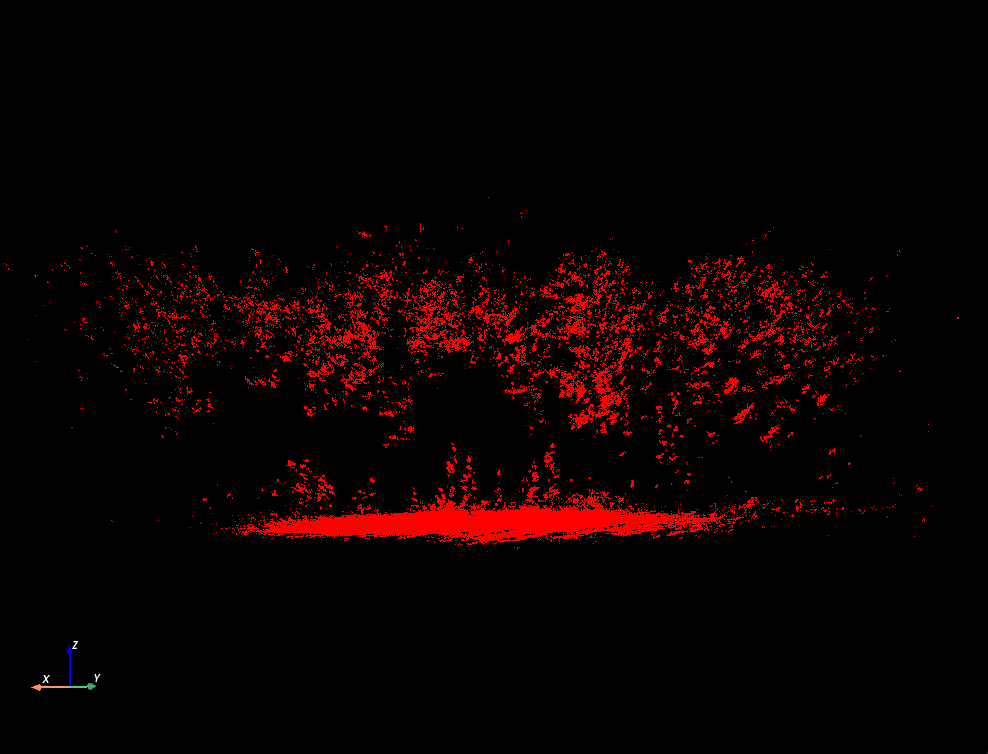}}
  \hspace{0.25mm}
	\subfloat[Complexity 3: $K=3$]{\includegraphics[width=0.3\linewidth]{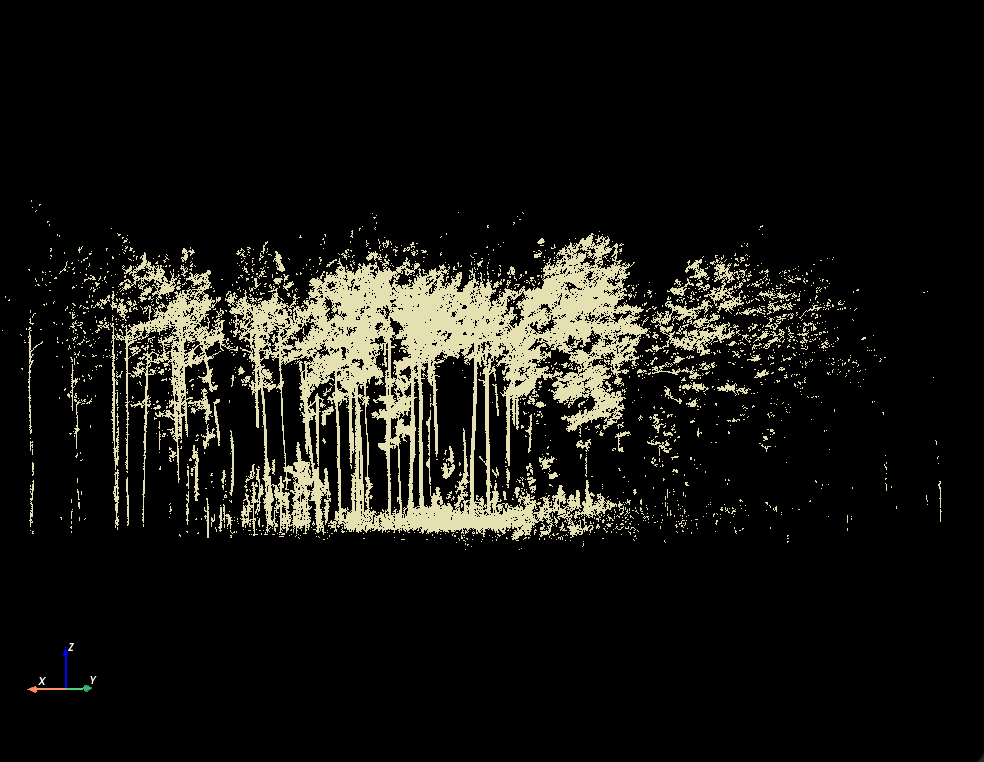}}
        \caption{3D points grouped by structural complexity using standard $k$-means do not naturally segment points into ground, tree trunks/branches, and foliage, unlike the vMF representation utilized in our approach.}
	\label{fig:splitting_comparisons} 
\end{figure}

Comparisons between the results of the mixture of vMF distributions in Fig. \ref{fig:structural_complexity} and the mixture of Gaussians for splitting point clouds and computing structural complexity are shown in Fig.\ref{fig:splitting_comparisons}. Note that the $k$-means approach was not able to split the point cloud into the underlying forest composition. This provides evidence that the mixture of Gaussians does not offer a natural clustering into structural complexities. The uneven sampling of 3D points within the range of the LiDAR sensor complicates the formation of fair groups of structural complexities. The standard $k$-means algorithm, assuming a mixture of Gaussians, results in clusters that reduce variance evenly as they assume independent and identically distributed 3D point samples. Overall, we qualitatively found that our structural complexity metrics and grouping approach splits point clouds from forests by their underlying components (e.g., ground, tree trunks/branches, foliage). 
%without requiring prior knowledge about their shape, as is typically assumed by standard segmentation approaches, nor the use of learning-based methods, which typically require many labeled training example point clouds, which can become impractical. 

One drawback of our method that we would like to point out is that the number of $K$-groups to which we categorize 3D points into structural complexity is generally unknown and can vary depending on the characteristics of the natural ecosystem. In our experiments so far, we used $K=2$ for ALS and $K=3$ for TLS throughout all cases, which resulted in good alignment performance overall. However, the number of groups $K$ is a parameter that has to be tuned for performance by trial and error. In the following, we present some additional co-registration experiments in ecosystems where there is no structure from trees and where a $K=2$ for TLS made more sense.

\subsubsection{Ecosystems without trees.}
\begin{figure}[ht]
	\centering 
	\subfloat[Side view]{\includegraphics[width=0.277\linewidth]{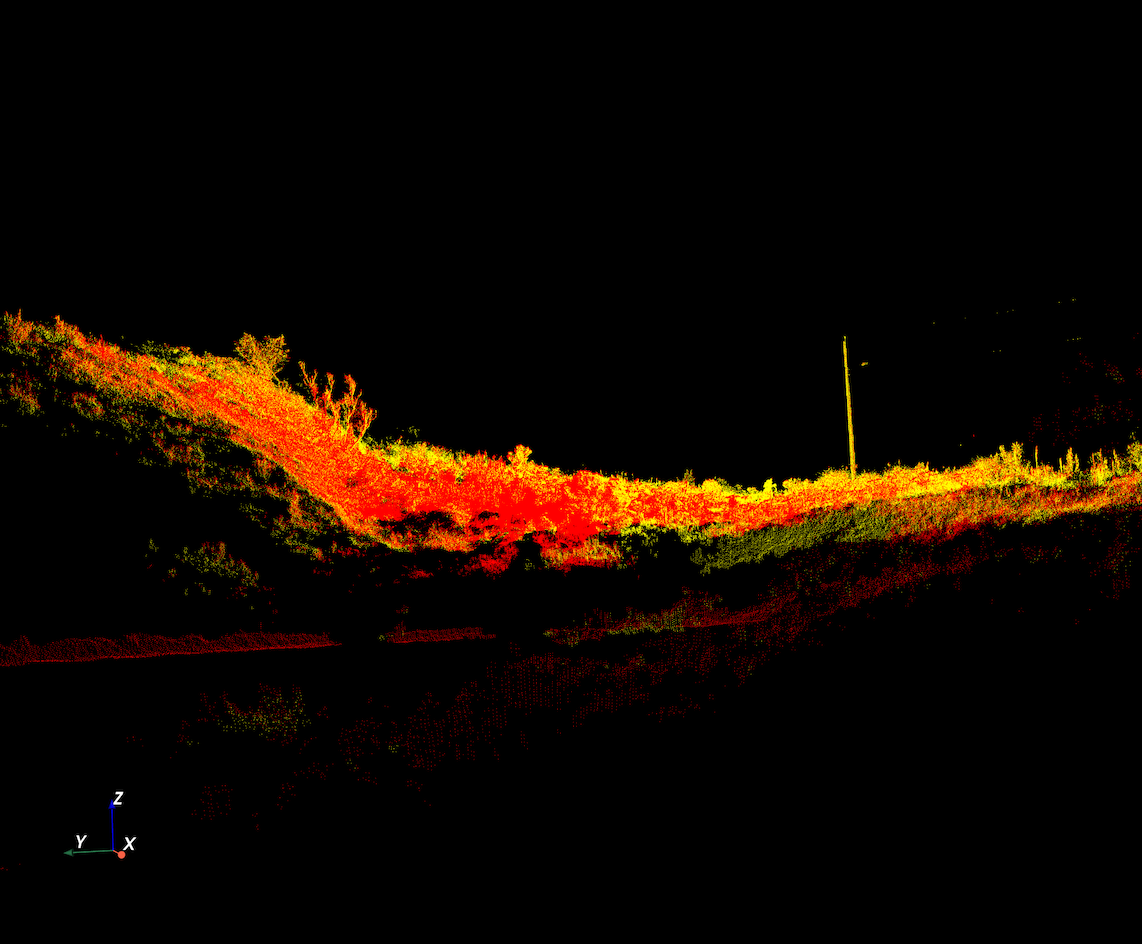}}
	\hspace{1mm}
	\subfloat[Ortho side view]{\includegraphics[width=0.26\linewidth]{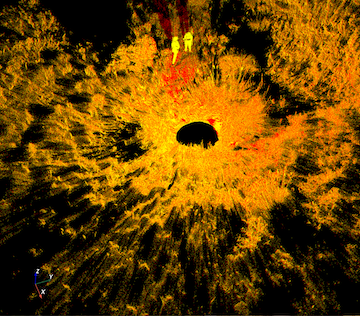}}
	\hspace{1mm}
	\subfloat[Side view]{\includegraphics[width=0.262\linewidth]{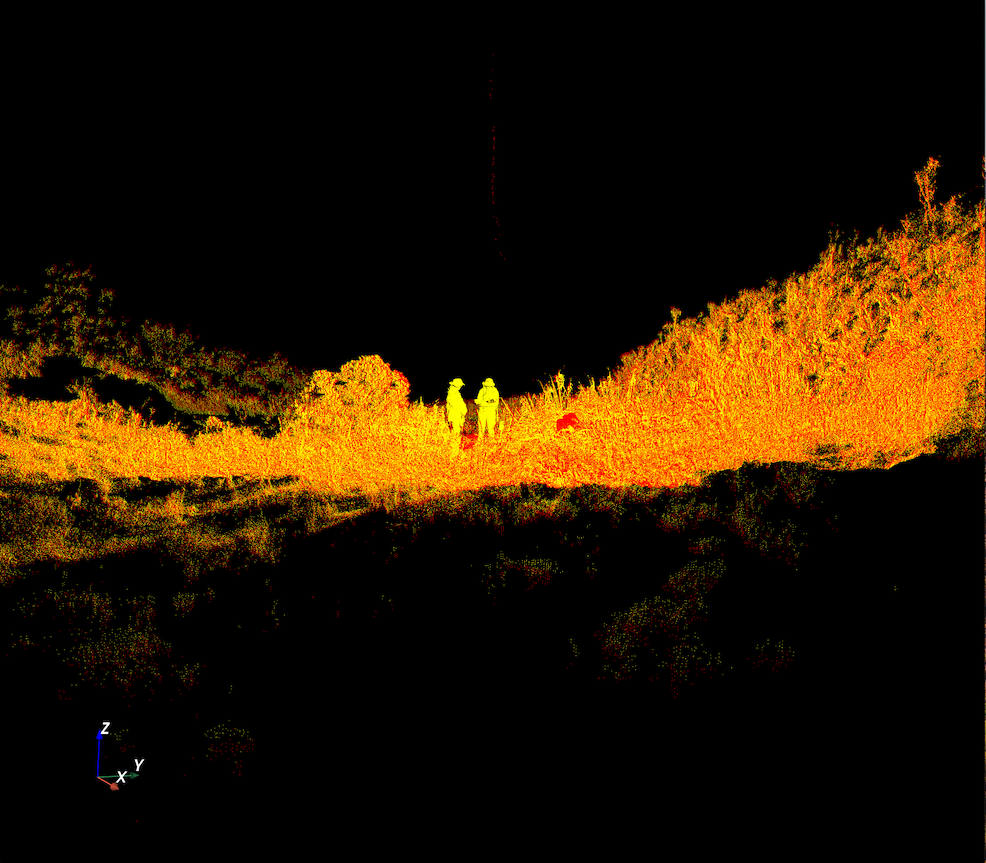}}
	%\subfloat[Post treatment point cloud]{\includegraphics[width=0.25\linewidth]{figures/prepost2}}
	\caption{Co-registration representative result in ecosystems without no trees. Source scan is color coded yellow while target scan is red. Note that our co-registration approach is effective in aligning point-clouds from grass-like ecosystems.}
	\label{fig:ecuador} 
\end{figure}
Here, we add a few additional evaluations demonstrating the capabilities of our co-registration approach to align point clouds in cases where the surveyed ecosystem does not contain trees. This aims to show the method's ability to exploit not only structural similarities between objects above the ground surface but also the surface topography itself. The dataset in this case is from the Paramos high-elevation grasslands in Ecuador. Fig.\ref{fig:ecuador} shows a representative example with pairs of source and target point clouds, each color-coded yellow and red, respectively.
In this case, we used $K=2$ structural complexity groups. We found that the 3D points were grouped mainly into the components of ground and grass, and our co-registration method resulted in well-aligned point clouds using this number. We note that this is an additional benefit of our proposed method: the ability to align point clouds in cases with no trees, where semantic feature-based methods (e.g., tree segmentation) can fail significantly, as there are no trees present to exploit.

\subsubsection{pre-TLS to post-TLS fire treatment effects registration}
\begin{figure}[ht]
	\centering 
	\subfloat[Pre-treatment RGB]{\includegraphics[width=0.17\linewidth]{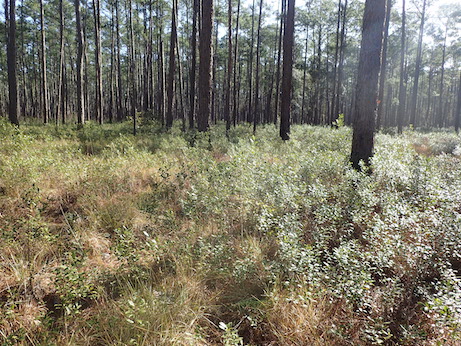}}
	\hspace{0.5mm}
	\subfloat[Post-treatment RGB]{\includegraphics[width=0.17\linewidth]{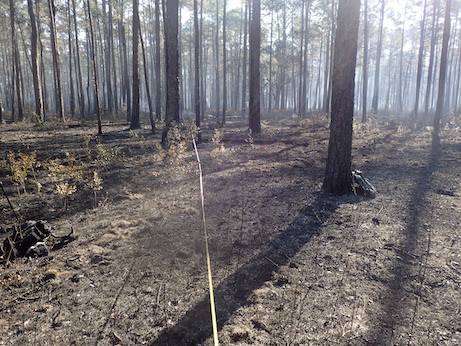}}
	\hspace{0.5mm}
	\subfloat[Ortho-side view]{\includegraphics[width=0.152\linewidth]{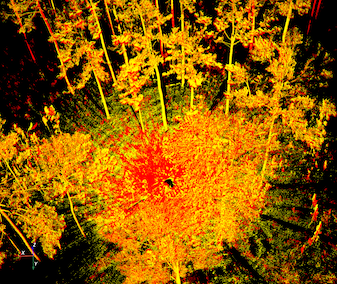}}
	\hspace{0.5mm}
	\subfloat[Side view]{\includegraphics[width=0.152\linewidth]{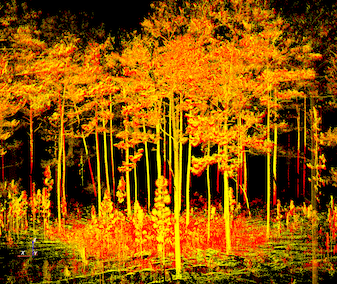}}
	\hspace{0.5mm}
	\subfloat[Side view]{\includegraphics[width=0.152\linewidth]{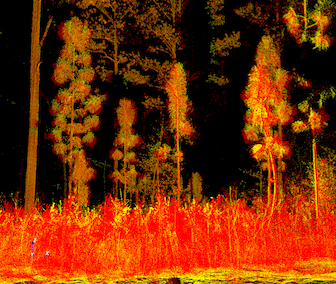}}
	\hspace{0.5mm}
	\subfloat[Side view]{\includegraphics[width=0.151\linewidth]{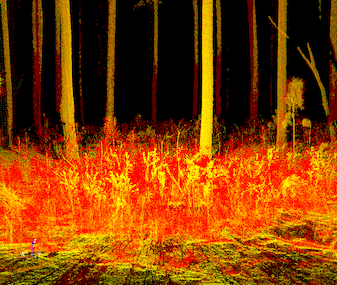}}
	\caption{Co-registration results in the presence of pre/post fire treatment effects.}
	\label{fig:prepost} 
\end{figure}
Evaluations in this case test the capabilities of pair-wise co-registration in forest ecosystems that underwent changes due to the effects of low-intensity fires targeting the ground surface. The scenes before and after fire treatments are referred to as the pre- and post-treatment point clouds, respectively. The severity of the fire was low, and the effects were qualitatively observable with morphologically thinning changes in the structure of surface fuels (e.g., grass, bushes). These pre- and post-fire scans were from a prescribed burn at Ft. Stewart Georgia, USA. For reference, we have included in Fig.\ref{fig:prepost}(a-b) RGB photos of pre/post-treatment effects. The structural complexity settings in this case set the number of structural complexity groups to $K=3$ in the pre-fire scan and $K=3$ in the post-fire scan. Representative results after applying our co-registration approach are included in Fig.\ref{fig:prepost}.(c-f), where we see effective alignment between the pre- and post-treatment point clouds under the given fire treatment effects. The pre-treatment point cloud is color-coded red, while the post-treatment point cloud is yellow. The resulting overall inlier RMSE was 0.014 m. Accurate alignment methods for point clouds experiencing pre/post fire or other scene changes can be of high value to science and forest managers. They enable analysis of treatment effects over time and at high-resolution spatial scales, a much-needed tool for better understanding prescribed fire behavior, which can ultimately lead to improved burn plans.

In terms of robustness to plausible erroneous initial estimates, we would like to highlight that our method is local, meaning it was designed to co-register point clouds miss-aligned by deviations less than 45° in rotations and a few meters in translations. Under these conditions the directional ambiguity often encountered when co-registering two planes (e.g., grounds) using standard approaches, such as ICP, does not arise. In our case, no issues were observed during the first stage of co-registration, where the lowest complexity features (typically the ground) were used to obtain a rough estimate of the co-registration parameters. Moreover, our approach assumes natural forests exhibit some variability in topography. Even in the Eastern US test dataset at Ft. Stewart, Georgia, where the terrain is known to be practically flat, our approach successfully detected subtle variability in the ground to obtain an initial co-registration estimate. Finally, the incremental nature of our approach, which aggregates points sequentially as complexity increases has a built-in capability to recover from erroneous initial estimates. Thus, if the ground were flat and an erroneous estimate were to be made at the lower complexity level, the method can recover at the next complexity level, where additional points above the ground can help to further constrain all six degrees of freedom characterizing the alignment between point clouds.

In the case where the ground is completely flat and lacks any vertical variability (e.g., uniform ground and grass), and without a good initial estimate of the co-registration parameters, our method would be able to align up to the $x,y$ dimensions due to the lack of constraints on all six degrees of freedom. However, we would like to contend that such scenarios are highly unlikely in real-world forest settings. 
In dense forest scenarios, the robustness of our approach depends on the structural diversity of the landscape. If sufficient topographic variability exists, the method can leverage these subtleties to constrain all six degrees of freedom and achieve a good initial alignment, which can then be further refined, similar to the Ecuador case, as long as the search space for correspondences remains within a close neighborhood. This also applies to dense foliage, regardless of understory structure. However, in the unlikely case of a completely flat ground with no variability to constrain all six degrees of freedom and dense, uniformly spaced vegetation (i.e., uniform tree spacing and foliage distribution), our approach may encounter difficulties. This challenge arises because the initial stages may not provide a good initial alignment, and subsequent, higher-complexity stages may fail to refine or lock the co-registration parameters at a global maximum. Instead, the dense foliage or uniform tree spacing may create numerous local minima in the optimization function, resulting in likely incorrect alignments. In intermediate scenarios, where natural forests exhibit some variability in topography and/or in the tree-stand direction, no issues were observed with co-registration.

In summary, tree-based methods perform well in some cases; however, their reliance on sparse tree-based features derived from the point cloud, rather than the entire point cloud, makes their accuracy highly dependent on the pre-processing stage used to estimate these tree-based features. This line of approaches sacrifice the level of spatial detail and feature richness of the point cloud, limiting the overall accuracy of the co-registration method. Moreover, increased occlusion in the scene generally leads to larger errors in tree feature predictions, which, in turn, introduce errors in the co-registration optimization. This is problematic in both TLS-to-TLS and TLS-to-ALS scenarios, in the former case as occlusion increasingly affects the level of information as a function of range, and in the later because of the LiDAR's limited ability to capture beyond the canopy. In contrast, our method leverages the full information available in the point cloud, similar in spirit to ICP but better suited for scenes with varying levels of structural complexity. Furthermore, our approach is capable of co-registering natural scenes even in the absence of trees (e.g., in prairies), as long as there is at least some subtle elevation variability in the topography. This is because our method can use ground variability to find rough alignments, unlike tree-based methods, which typically do not exploit ground features for co-registration. 
Moreover, our approach can exploit different levels of structural complexity to achieve co-registration in both TLS-to-TLS and TLS-to-ALS scenarios. In the TLS-to-TLS case, it implicitly leverages the advantages of trunk-based methods by initially focusing on under-canopy features to obtain a rough alignment. In the TLS-to-ALS case, it starts by focusing on ground variability features to achieve a rough alignment, and then refines the alignment using features around the canopy level. This is an advantage that tree-feature based methods do not provide, and rather, one has to use different co-registration methods on a case by case basis. Overall, our approach outperformed forest-semantic-based co-registration methods and achieved comparable results to manual co-registration and target-based methods, with the added advantage of being fully automatic and target-free.

One additional drawback of the proposed method is that it takes approximately 5 minutes to complete a single TLS pair-wise co-registration. In the case of TLS-to-ALS co-registration it was approximately 3 minutes. This may not represent a problem when only co-registering a few pairs of point clouds, as is typically the case with existing standard forest surveying practices. However, scaling to thousands or even millions of point cloud pairs can become impractical with the current implementation. To this end, improvements to computational efficiency could be explored by taking advantage of parallelization strategies, which could significantly reduce the convergence time of ForestAlign.

In terms of additional benefits, we would like to point out that our innovation enables the capability to produce integrated LiDAR maps automatically with the attractive flexibility of access to broad low-res ALS data to rapidly query ecosystem attribute estimation summaries and at the same time access to detailed co-registered high-resolution in-situ understory TLS data for more detailed refined inference of characterizations at the available sites. This flexibility offers the potential to exploit and demonstrate the advantages of either individual or integrated LiDAR sensing sources for ecosystem analysis while also opens up avenues to define the necessary conditions and best strategies for more efficient sensing in ecosystem monitoring. 
%
%Recent work by \cite{pokswinski2021simplified, anderson2021traditional, loudermilk2023terrestrial} has demonstrated the efficiency and efficacy of ecosystem monitoring using single scan in-situ TLS. The technological advances of such models include new capabilities for rapidly extracting highly detailed quantifiable predictions of vegetation attributes and treatment effects in near surface, mid- and general under-story composition. However, these models have only been deployed across spatial domains of a few tens of meters in radius. Our innovation here integrating multiple TLS sources and ALS  and their synergistic use has the ability to scale in-situ TLS findings as in \cite{pokswinski2021simplified} to deployments over larger area extents. 
%
The proposed co-registration approach is an enabling technology for large scale ecosystem monitoring through more advanced computational tools of data source integration while also offering new possibilities to more advanced analysis at scale. It also enables the ability to monitor sites with a more efficient framework to update ecosystem information (as in the pre/post Rx fire co-registation) and with this, the capability to track localized changes over time. We think our co-registration methodology can also help in establishing trade-offs between ALS, TLS and integrated ALS+TLS based monitoring comparisons and relative to standard forest plot surveying. This, measured trough comparisons by time efficiency, human labor requirements, associated costs, performance accuracy, scalability. In future work, we intend to explore a comprehensive investigation of these tradeoffs through the aforementioned metrics. These capabilities altogether can overall improve planning, management and more efficient decision-making of natural ecosystem lands. %For example by providing quick and cheap characterizations or forest structure essential for wildlife habitat protection and prescribed burn planning applications.

\section{Conclusion}
\label{Sec:conclusion}

In this work, we proposed an automatic targetless multi structural complexity level approach to register pair-wise multi-view TLS LiDAR scans and ALS scans to a common reference coordinate system in complex heterogeneous forests. 
%The resulting alignments were further refined using a bundle refinement procedure which promotes cycle consistency between the point clouds to further refine their alignments into a consistent global coordinate system. 
The qualitative and quantitative results of our experimentation validate that the proposed approach is both efficient and effective in aligning LiDAR scans in forestry at performances comparable with target based methods, but without any target placement requirements. It, in addition, showed to be robust in a variety of cases including  a diverse set of resolution scales, view-points, overlap, forest heterogeneity (e.g., shrubs below the canopy, different size trees, steepness of the ground surface), low-intensity fire effects  and random alignment initializations.

\section*{Acknowledgement}
Research presented in this article was supported by the Laboratory Directed Research and Development program of Los Alamos National Laboratory under project number 20220024DR. We thank the United States Department of Agriculture (USDA) Forest Service, Southern Research Station for their support. This research was funded in part by the Department of Defense, Strategic Environmental and Research Development Program, grants number RC19-1119 and RC20-1346, and Department of Defense, Environmental Security Technology Certification Program, grant number RC20-7189. This research was also funded in part by the USDA Center for International Programs and U.S. Agency for International Development. The findings and conclusions in this publication are those of the authors and should not be construed to represent any official USDA or U.S. Government determination or policy.

%%%%%%%%%%%%%%%%%%%%%%%%%%%%%%%%%%%%%%%%%%%%%
%% Bibliography
%%%%%%%%%%%%%%%%%%%%%%%%%%%%%%%%%%%%%%%%%%%%%

\bibliographystyle{elsarticle-num}
%\bibliography{refs_clean.bib}

\end{document}